\documentclass[10pt,journal, compsoc]{IEEEtran}
\usepackage{amsmath,amsfonts}
\usepackage{algorithm}
\usepackage{algpseudocode}
\usepackage{booktabs}
\usepackage{array}
\usepackage[caption=false,font=normalsize,labelfont=sf,textfont=sf]{subfig}
\usepackage{textcomp}
\usepackage{stfloats}
\usepackage{url}
\usepackage{verbatim}
\usepackage{graphicx}
\usepackage{xcolor}
\usepackage[numbers,square,sort]{natbib}
\usepackage{makecell}

\definecolor{mycolor}{HTML}{000000}
\usepackage[
        breaklinks,
        colorlinks,
        citecolor=blue,
        linkcolor=blue,
        ]{hyperref}
\usepackage{multirow}

\begin{document}

\title{{{AgentsCoMerge}}: Large Language Model Empowered Collaborative Decision Making for Ramp Merging}

\author{Senkang Hu,~\IEEEmembership{Graduate Student Member,~IEEE},  Zhengru Fang,~\IEEEmembership{Graduate Student Member,~IEEE},\\
Zihan Fang,~\IEEEmembership{Graduate Student Member,~IEEE},  Yiqin Deng,~\IEEEmembership{Member,~IEEE},\\Xianhao Chen,~\IEEEmembership{Member,~IEEE}, Yuguang Fang,~\IEEEmembership{Fellow,~IEEE}, Sam Tak Wu Kwong,~\IEEEmembership{Fellow,~IEEE} 
\IEEEcompsocitemizethanks{\IEEEcompsocthanksitem Senkang Hu, Zhengru Fang, Zihan Fang, Yiqin Deng, and Yuguang Fang are with the Hong Kong JC STEM Lab of Smart City and the Department of Computer Science, City University of Hong Kong, Kowloon, Hong Kong. (Email: \{senkang.forest, zhefang4-c, zihanfang3-c\}@my.cityu.edu.hk, \{yiqideng, my.Fang\}@cityu.edu.hk) 
\IEEEcompsocthanksitem Xianhao Chen is with Department of Electrical and Electronic Engineering, The University of Hong Kong, Pok Fu Lam, Hong Kong. (Email: xchen@eee.hku.hk)
\IEEEcompsocthanksitem Sam Tak Wu Kwong is with the Department of Computing and Decision Sciences, Lingnan University, Hong Kong. (Email: samkwong@ln.edu.hk)
}
\thanks{The research work described in this paper was conducted in the JC STEM Lab of Smart City funded by The Hong Kong Jockey Club Charities Trust under Contract 2023-0108.  The work was supported in part by the Hong Kong SAR Government under the Global STEM Professorship and Research Talent Hub. The work of S. Hu was  supported in part by the Hong Kong Innovation and Technology Commission under InnoHK Project CIMDA. The work of Y. Deng was supported in part by the National Natural Science Foundation of China under Grant No. 62301300. The work of X. Chen was supported in part by HKU-SCF FinTech Academy R\&D Funding. (Corresponding author: Yiqin Deng)}
}

\IEEEtitleabstractindextext{%
\begin{abstract}
Ramp merging is one of the bottlenecks in traffic systems, which commonly cause traffic congestion, accidents, and severe carbon emissions. In order to address this essential issue and enhance the safety and efficiency of connected and autonomous vehicles (CAVs) at multi-lane merging zones, we propose a novel collaborative decision-making framework, named \textbf{AgentsCoMerge}, to leverage large language models (LLMs). Specifically, we first design a scene observation and understanding module to allow an agent to capture the traffic environment. Then we propose a hierarchical planning module to enable the agent to make decisions and plan trajectories based on the observation and the agent's own state. In addition, in order to facilitate collaboration among multiple agents, we introduce a communication module to enable the surrounding agents to exchange necessary information and coordinate their actions. Finally, we develop a reinforcement reflection guided training paradigm to further enhance the decision-making capability of the framework. Extensive experiments are conducted to evaluate the performance of our proposed method, demonstrating its superior efficiency and effectiveness for multi-agent collaborative decision-making under various ramp merging scenarios. 
\end{abstract}

\begin{IEEEkeywords}
  Collaborative Decision Making, Large Language Model (LLM), Connected and Autonomous Vehicle (CAV), Multi-Lane Merging.
\end{IEEEkeywords}}

\maketitle

\IEEEdisplaynontitleabstractindextext

%
\IEEEpeerreviewmaketitle

\section{Introduction}

\IEEEPARstart{L}{ane} merging areas at ramps, are commonly recognized as the bottlenecks within highway system, that can significantly disrupt the flow of traffic on the main roadways \cite{karbalaiealiDynamicAdaptiveAlgorithm2020,zhuMergingControlStrategies2022}. Specifically, in a lane merging maneuver, a vehicle in the ramp has to merge into the main road within a restricted merging area. For those navigating from a ramp, determining whether to speed up or slow down to safely enter a busy main road can be highly challenging. This decision-making process is particularly difficult for elderly or inexperienced drivers \cite{dewaardMergingHeavyMotorway2009}. In heavy traffic conditions, it might even be necessary for a vehicle to halt completely before merging. Entering the main road at a low speed or with inadequate space can cause the vehicles on the main road to slow down, either abruptly or voluntarily, which often results in various adverse effects on both traffic efficiency and safety. As a result, the development of effective strategies to enhance the safety and efficiency of CAVs at multi-lane merging zones is of paramount importance.

Recently, with the development of autonomous driving (AD) technology which offers a safer and more efficient driving system, the research community has made significant progress in addressing the collaborative decision-making problem for ramp merging. Most existing decision-making systems from industry and academia use rule-based algorithms such as Finite State Machines,  probability-based methods such as Markov Decision Process, or learning-based methods such as deep learning and reinforcement learning (RL) \cite{wangDecisionMakingDrivenDriver2023}. 
However, these methods inevitably face a lot of challenges when applied to real-world scenarios. For example:
\begin{enumerate}
    \item {Diversity of drivers' intentions}: The characteristics and needs of human drivers or passengers are complex and vary independently; therefore, adapting to human requirements is essential. However, established methods are often unable to fulfill these demands. 
    \item {Lack of interpretability}: Informational interpretability is essential for humans to ensure safety and foster trust in autonomous driving systems. However, traditional approaches often fall short in terms of interpretability, which remains a consistent issue in existing autonomous driving systems. This problem arises because these conventional methods typically rely on black-box deep neural networks, making it challenging to comprehend the decision-making process, reasoning, and logical thinking underlying their decisions. On the other hand, while rule-based and probability-based models offer high interpretability, they struggle to deliver adequate performance across a wide range of scenarios. So it is challenging for a system to realize both high interpretability and high performance.
    \item {Manual discretization}: Manual discretization involves transforming a continuous intermediate variable into a finite number of states to simplify the modeling process. This technique is frequently employed in certain rule-based and RL-based methods. Despite its role in simplifying the modeling process, discretization can significantly distort the model's reflection of the real world. In actual driving scenarios, drivers constantly face dynamic and continuous environments. Therefore, models with less discretization more accurately mirror real-world driving conditions.
\end{enumerate}


In order to overcome the aforementioned limitations, a powerful framework that can handle the divers' different intentions, provide high interpretability, and avoid overmuch manual discretization at the same time is needed. It was impossible to design this kind of framework a few years ago. However, luckily, large language models (LLMs) have emerged as a revolutionary technology in artificial intelligence (AI) and have been recognized as the early stages of artificial general intelligence (AGI) \cite{bubeck2023sparks}, which show great capabilities that closely mimic human thought processes. They can reason through novel situations using inherent commonsense knowledge, learn from previous experiences, and offer insights into their decision-making processes. These qualities position LLMs as a potent tool for addressing existing challenges and creating a robust framework for collaborative decision-making in ramp-merging.

In this paper, we leverage LLMs as the core component to build a novel framework,        {AgentsCoMerge}. This framework consists of four parts. First, we introduce a scene understanding module, which can not only understand text information but also process visual information. Following the scene understanding module, a hierarchical planning module is introduced to generate the driving decision and plan the trajectory. In addition, to facilitate the agent's ability to communicate with others and make collaborative decisions, an inter-agent communication module is proposed. Finally, we design a reinforcement reflection guided training strategy to further enhance the agents' decision-making capabilities. 
Our contributions are summarized as follows.
\begin{enumerate}
    \item We propose a novel framework, {AgentsCoMerge}, which consists of observation, cognitive memory, reasoning engine, iterative reinforcement reflection, and communication modules, to enable multi-vehicle collaborative driving.
    \item We are the first to leverage LLMs to enable collaborative decision-making tailored for multi-lane merging scenarios in autonomous driving and intelligent transportation systems.
    \item We conduct extensive experiments to evaluate the performance of our proposed method, demonstrating its superior efficiency and effectiveness in multi-agent collaborative decision-making under various lane merging scenarios.
\end{enumerate}

The remainder of this paper is organized as follows. In Section \ref{sec:related_work}, we review the work related to the topic of this paper, including large language models, vision language models, language models in autonomous driving, multi-agent collaborative decision-making, and ramp merging in autonomous driving. In Section \ref{sec:system_model}, we formulate the situation model, vehicle model, and multi-agent collaborative decision-making model. In Section \ref{sec:method}, we present the proposed {AgentsCoMerge} framework in detail. In Section \ref{sec:experiments}, we conduct experiments to evaluate the performance of our proposed method. Finally, we conclude the paper in Section \ref{sec:conclusion}.

\section{Related Work}
\label{sec:related_work}

\subsection{Large Language Model and Vision Language Model}

A Large Language Model (LLM) is a type of artificial intelligence model that specializes in processing and generating human-like text from the input they receive \cite{brownLanguageModelsAre2020, huAgentsCoDriverLargeLanguage2024,linPushingLargeLanguage2023, fangAutomatedFederatedPipeline2024}. LLMs are mostly based on Transformer architecture, which was introduced by Vaswani \textit{et al.} \cite{vaswaniAttentionAllYou2017}. They are characterized by their vast number of parameters, which are essentially the aspects of the model that are learned from data and determine the model's behavior. Models like GPT-3 \cite{brown2020language}, GPT-4 \cite{sanderson2023gpt}, PaLM \cite{chowdhery2023palm}, and LLaMA \cite{touvronLLaMAOpenEfficient2023}, for example, have billions to hundreds of billions of parameters, allowing them to capture a wide range of language nuances and complexities. In addition, other open-sourced LLMs such as GLM \cite{duGLMGeneralLanguage2022} and OPT \cite{zhangOPTOpenPretrained2022} have also achieved high performance, and they serve as crucial software resources for training large models and as a basis for additional fine-tuning for specific applications. For example, Alpaca \cite{alpaca} presents a self-instruct framework that enhances instruction tuning for a LLaMA model, decreasing the dependence on human-generated instruction data. Vision language models (VLMs) are multimodal LLMs, which aim to handle visual inputs, such as images and videos. Li \textit{et al.} \cite{liBLIP2BootstrappingLanguageImage2023} proposed BLIP to leverage Q-formers to align an image input with a text input, enabling the model to understand images and generate descriptions. Li \textit{et al.} \cite{liLLaVAMedTrainingLarge2023} directly leveraged a fully connected neural network to project image tokens to the text space. VLMs have been used in various applications, such as image understanding, video captioning, medical diagnosis, and robotics. In the context of autonomous driving, VLMs have also shown great potential in understanding the driving scenarios and facilitating decision-making processes.

\subsection{Language Models in AD}

The progress in LLMs and VLMs has demonstrated impressive human-like capabilities, including understanding, reasoning, and zero-shot learning. These features position LLMs as a promising solution to AD in areas such as decision-making, scenario understanding, reasoning, and planning. For instance, Cui \textit{et al.} \cite{cuiDriveYouSpeak2023} utilized the linguistic and reasoning capabilities of LLMs in autonomous vehicles to implement an interactive design through human-machine dialogues. Similarly, Wen \textit{et al.} \cite{wenDiLuKnowledgeDrivenApproach2023} introduced a knowledge-driven approach using LLMs to incorporate memory and reflection modules into AD, thereby improving the decision-making process.
Additionally, VLMs are capable of processing images and video data to understand scenarios, connecting these visual inputs with understandable texts to facilitate comprehensive perception in decision-making processes. For example, Pan \textit{et al.} \cite{panVLPVisionLanguage2024} presented VLP, a vision-language-planning framework, to leverage language models to seamlessly integrate linguistic comprehension with the requirements of AD, which can strengthen the source memory foundation and the autonomous vehicles' contextual understanding. Han \textit{et al.} \cite{hanDMEDriverIntegratingHuman2024} proposed DME-Driver to integrate the pre-annotated GPT-4V(ision) system with human-like logic through manual adjustments, harmonizing logical decision-making with accurate control signals for AD. It also developed specialized models by utilizing the LLaVA framework for enhanced training. Sima \textit{et al.} \cite{simaDriveLMDrivingGraph2023} developed DriveLM, a new baseline for end-to-end AD. DriveLM presents graph visual question answering to organize the question-answer pairs into a graph structure via logical dependencies at the object level. 

\subsection{Multi-Agent Collaborative Decision-Making}

In the domain of multi-agent collaboration and communication, there has been a notable interest in utilizing LLMs for enhancing cooperative efforts \cite{huAdaptiveCommunicationsCollaborative2023,10779389,huCollaborativePerceptionConnected2024,huAgentsCoDriverLargeLanguage2024,fangPACPPriorityAwareCollaborative2024}. 
\textcolor{black}{Zhang \textit{et al.} \cite{zhangBuildingCooperativeEmbodied2023} designed a system utilizing LLMs to enhance cooperative behavior among embodied agents, facilitating effective communication to achieve common objectives in diverse settings. Hong \textit{et al.} \cite{hongMetaGPTMetaProgramming2023} created MetaGPT to incorporate human involvement in LLM-guided collaborative activities, thus combining human intuition with the capabilities of artificial intelligence. Furthermore, Li \textit{et al.} \cite{liCAMELCommunicativeAgents2023} introduced CAMEL, a system that establishes role-based agent interactions, fostering autonomous collaboration through dynamic role engagement.}
\textcolor{black}{Specifically, in the realm of connected and autonomous vehicles (CAVs), the emphasis on collaborative decision-making utilizing LLMs has intensified in accident analysis, scene simulation, and intersection management. For instance, Wang \textit{et al.} \cite{wang_accidentgpt_2023} developed AccidentGPT, an LLM-based framework integrating multi-sensor data to facilitate comprehensive accident analysis and prevention for traffic safety. Most recently, Hu \textit{et al.} \cite{huAgentsCoDriverLargeLanguage2024} developed a collaborative driving framework employing LLMs to accumulate knowledge and experience for lifelong learning, which demonstrated its superior communication and collaboration efficiency in highway and intersection scenarios.}
However, in our paper, we focus on harnessing LLMs and VLMs to facilitate comprehensive collaborative decision-making under on-ramp merging scenarios. By leveraging the advanced capabilities of foundation models, we aim to develop a system to coordinate and manage decision-making processes among multiple agents under multi-lane merging scenarios, thereby advancing the safety, efficiency, and reliability of connected and autonomous driving (CAD).

\subsection{Ramp Merging in AD}

Ramp merging plays a crucial role in modern transportation systems, as the actions of vehicles entering from ramps often introduce continuous disruptions to the flow of traffic on the main roads. Such disturbances can lead to a range of issues, such as fluctuations in traffic flow, higher fuel consumption and carbon emissions, safety risks, and regular occurrences of traffic jams \cite{zhuMergingControlStrategies2022,10138695,yangMultiLaneCoordinatedControl2023,10449899,10529950,10.1109/TMC.2024.3359040}. In order to tackle this issue, many methods have been proposed. For example, Zhou \textit{et al.} \cite{zhouImpactCooperativeAutonomous2016} introduced a controller in on-ramp areas, utilizing a collaborative intelligent driver model to address issues related to traffic oscillations and merging efficiency. Chen \textit{et al.} \cite{chenConnectedAutomatedVehicle2021} proposed a rotation-based CAV distributed collaborative control mechanism, which reduces the complexity and dimension of the problem by transferring the merging problem to a virtual car following problem. Moreover, there are also many optimization-based methods. For example, in Karbalaieali's work \cite{karbalaiealiDynamicAdaptiveAlgorithm2020}, the overall travel delay experienced by vehicles on an on-ramp under various traffic conditions was minimized by optimizing a cost function that considered the travel times of both ramp and main traffic flow. Liu \textit{et al.} \cite{liuEfficientOnRampMerging2022} also examined the variability in traffic flow across lanes and developed a model for lane selection to facilitate efficient merging and selection processes for vehicles on a ramp. Although the aforementioned methods have made significant contributions to the field, the integration of LLMs into the decision-making process for highway merging in AD has not been fully explored. In this paper, we propose a novel approach that leverages LLMs to enable end-to-end collaborative decision-making for highway merging in AD.


\section{System Model}
\label{sec:system_model}

\subsection{Situation Model}

\begin{figure}[t]
    \centering
    \includegraphics[width=1\linewidth]{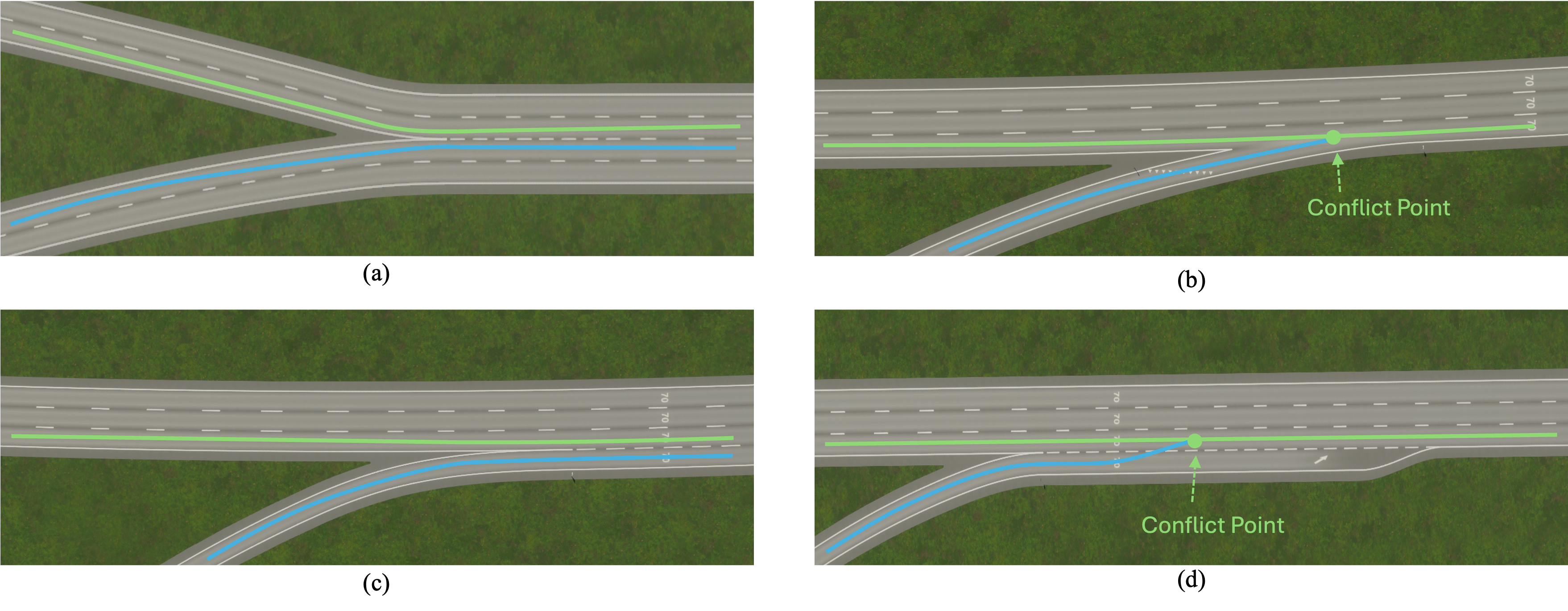}
    \vspace{-5mm}
    \caption{The multi-lane merging scenarios}
    \label{fig:multi_lane_merging_scenario}
    \vspace{-5mm}
\end{figure}

For multi-lane merging scenarios, as shown in Fig. \ref{fig:multi_lane_merging_scenario}, there are two merging conditions. The first condition is that there are no conflicting routes between the CAVs on a ramp and the main road, as shown in Fig. \ref{fig:multi_lane_merging_scenario}(a) and (c). In this case, prior to the merging point, the sum of the lanes on the main road and the ramp is equal to the total number of lanes at the merge point. Therefore, the CAVs in the ramp can merge into the main road without changing lanes.
The second condition is that there are conflicting routes between the CAVs in a ramp and the main road, as shown in Fig. \ref{fig:multi_lane_merging_scenario}(b) and (d). In this case, the sum of the lanes on the main road and the ramp is greater than the total number of lanes at the merge point. Therefore, the routes of CAVs in the ramp lane have conflicts with the routes of CAVs in the main road. In this paper, we only consider the second scenarios, because the first condition has no conflict point, so it can be simplified to a lane maintenance issue rather than a ramp merging problem. 

In addition, under multi-lane merging scenarios, we only consider the coordination of CAVs, which are equipped with V2X communication devices that can communicate with other CAVs, different sensors such as cameras, LiDAR, inertial measurement unit (IMU), and GPS that can perceive the surrounding environment and measure the vehicle's acceleration, angular velocity and position coordinates.

Under these situations, our goal is to develop a collaborative decision-making framework for CAVs to merge into the main road smoothly and safely.

\subsection{Vehicle Formulation}

Multi-lane merging is a dynamic and complex process where vehicle states change continuously over time. In order to accurately control the CAV to have a smooth and safe merging process, it is essential to model the vehicle dynamics so that we can have a better understanding of the CAV's motion and accurately capture the dynamic changes. 

\begin{figure}[t]
    \centering
    \includegraphics[width=0.8\linewidth]{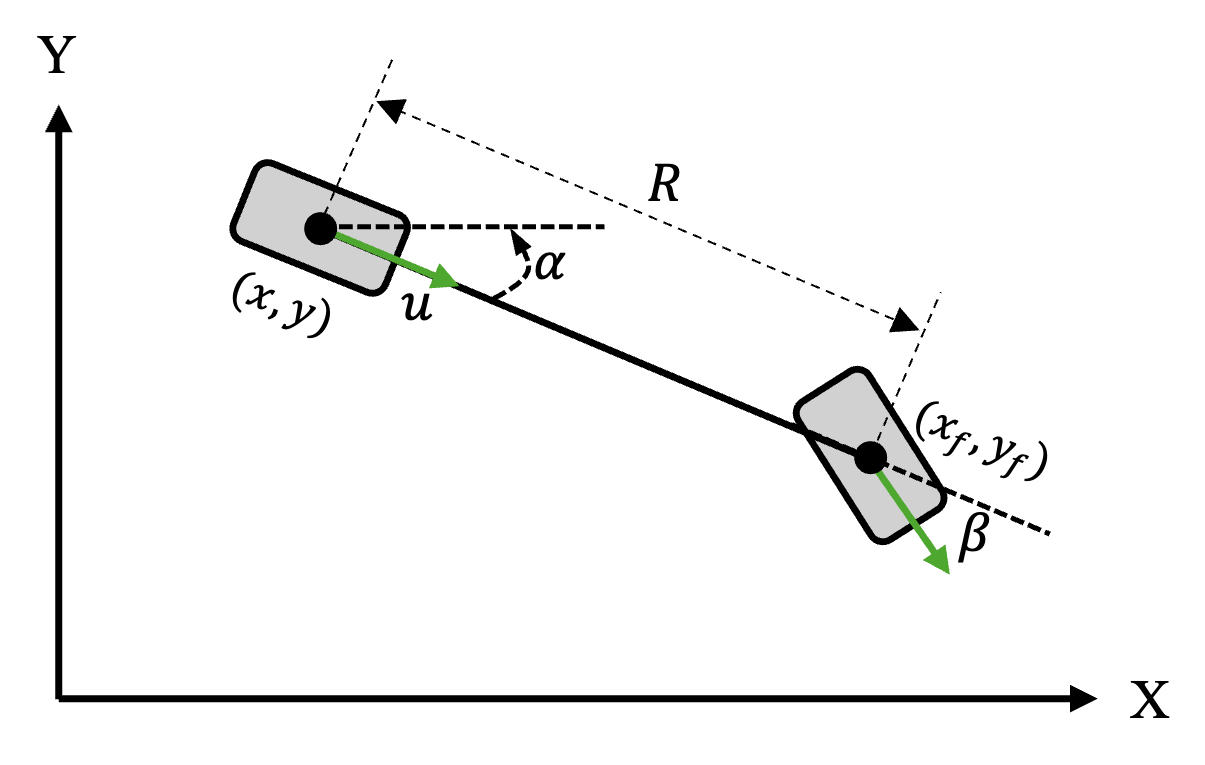}
    \vspace{-5mm}
    \caption{The kinematic bicycle model of a CAV}
    \label{fig:CAV_kinematic_model}
    \vspace{-5mm}
\end{figure}

Specifically, the CAV dynamics can be modeled using a kinematic bicycle model \cite{rajamaniVehicleDynamicsControl2006}, as illustrated in Fig. \ref{fig:CAV_kinematic_model}. The bicycle model treats a vehicle as if it had a single front and a single rear wheel, similar to a bicycle. This simplification reduces the four-wheel vehicle dynamics into a two-wheel model, focusing on lateral dynamics and ignoring complex interactions such as tire slip angles, lateral forces, and suspensions. 

In Fig. \ref{fig:CAV_kinematic_model}, $(x, y)$ are the coordinates of the rear wheel, $(x_f, y_f)$ are the coordinates of the front wheel, $\alpha$ is the orientation of the vehicle, and $\beta$ is the steering angle of the front wheel. The radius of the turning circle is denoted as $R$. $u$ is the longitudinal velocity of the vehicle and $\omega$ is the angular velocity of the steering wheel. Then, the state of the vehicle can be described as $\mathbf{S}= [x, y, \alpha, \beta]^T$, and $\mathbf{C} = [u, \omega]^T$ as the control input. In addition, the kinematic model is constrained by non-holonomic conditions because it assumes that the wheels roll without slipping on the ground. In this model, the orientation of the rear wheel is fixed while the front wheel can be steered. Therefore, we have 
\begin{align} \label{eq:kinematic_bicycle_model_1}
    \dot{x}_f &= u \cos \alpha, \\
    \label{eq:kinematic_bicycle_model_2}
    \dot{y}_f &= u \sin \alpha.
\end{align}
\textcolor{mycolor}{where $\dot{x}_f$ and $\dot{y}_f$ are the velocities of the front wheel. The overdot means the derivative of the variable with respect to time. In addition, the front wheel's velocity must align with its steering direction $\alpha+\beta$, the non-holonomic constraint for the front wheel is}
\begin{equation}
    \dot{x}_f\sin(\alpha+\beta) = \dot{y}_f\cos(\alpha+\beta).  
    \label{eq:kinematic_bicycle_model_3}
\end{equation}
With the  rigid-body constraint, we have
\begin{align}
    x_f &=      x + R\cos\alpha,\\
    y_f &= y + R \sin\alpha.
\end{align}
\textcolor{mycolor}{Differentiating \( x_f \) and \( y_f \) with respect to time, we obtain
\begin{align}
  \dot{x}_f &= \dot{x} - R \sin \alpha \cdot \dot{\alpha}, \\
  \dot{y}_f &= \dot{y} + R \cos \alpha \cdot \dot{\alpha}.
\end{align}
Substituting \( \dot{x} = u \cos \alpha \) and \( \dot{y} = u \sin \alpha \), we obtain
\begin{align}
  \dot{x}_f &= u \cos \alpha - R \sin \alpha \cdot \dot{\alpha}, \\
  \dot{y}_f &= u \sin \alpha + R \cos \alpha \cdot \dot{\alpha}.
\end{align}
Then, we can substitute $\dot{x}_f$ and $\dot{y}_f$ into Eq. (\ref{eq:kinematic_bicycle_model_3}), we have
\begin{align*}
&\left( u \cos \alpha - R \sin \alpha \cdot \dot{\alpha} \right) \sin(\alpha + \beta) =\\
&\left( u \sin \alpha + R \cos \alpha \cdot \dot{\alpha} \right) \cos(\alpha + \beta).
\end{align*}
Thus, we can expand and group terms involving \( u \) and \( \dot{\alpha} \) to arrive at
\begin{align*}   
u \Big[ \cos \alpha \sin(\alpha + \beta) - \sin \alpha \cos(\alpha + \beta) \Big] &- \\ 
R \dot{\alpha} \Big[ \sin \alpha \sin(\alpha + \beta) + \cos \alpha \cos(\alpha + \beta) \Big] & = 0,
\end{align*}
which can be simplified to
\begin{align}
u \sin \beta - R \dot{\alpha} \cos \beta = 0.
\end{align}
Therefore, we finally have 
}
\begin{equation}
    \dot{\alpha} = \frac{u}{R} \tan \beta. \label{eq:kinematic_bicycle_model_4} 
\end{equation}
Thus, the dynamics of this model $\dot{\mathbf{S}}= f_k(\mathbf{S}, \mathbf{C})$ can be described as
\begin{equation} 
\left\{\begin{aligned} 
        \dot{x} &= u \cos \alpha \\
        \dot{y} &= u \sin \alpha \\
        \dot{\alpha} &= \frac{u}{R} \tan \beta \\
        \dot{\beta} &= \omega
\end{aligned}\right.
\end{equation}

\textbf{Definition 1} \cite{lynchModernRoboticsMechanics2017}: \textit{
A system is termed \textbf{differentially flat} if there exists a special type of output (referred to as the flat output) that can express all the system's states and control inputs purely in terms of this output and its derivatives. No integration is needed to recover the states or controls from the flat output. Mathematically, a system with state $s\in R^n$ and control inputs $c\in R^m$ is differentially flat if there exists outputs $y \in  R^m$ of the form
$$y=h(s, c, \dot{c}, \ldots, c^{(a)}),$$
such that the states and control inputs can be expressed as
\begin{align}
    s&=\varphi(y,\dot{y}, \ldots, y^{(b)} ),\\
    c&=\gamma(y,\dot{y}, \ldots, y^{(b)} ).
\end{align}
}

The kinematic bicycle model has been demonstrated to exhibit differential flatness that is defined in Definition 1. Specifically, when we select $\delta = (\delta_1, \delta_2) = (x, y)$ as the flat outputs, owing to this attribute of differential flatness, it is possible to express all the states of a vehicle solely in terms of the flat outputs and a finite number of their derivatives. This capability significantly aids in the formulation of the vehicle's trajectory and enhances the process of motion planning \cite{van1998differential}. We demonstrate this attribute as follows. 

Firstly, we can obtain the longitudinal velocity $u$ by solving Eq. (\ref{eq:kinematic_bicycle_model_1}) and (\ref{eq:kinematic_bicycle_model_2}), which can be expressed as
\begin{equation}
    u=\sqrt{\dot{\delta_1}^2 + \dot{\delta_2}^2}.
\end{equation}
The orientation of the vehicle can be expressed as
\begin{equation}
    \alpha = \arctan({\dot{\delta_2}}/{\dot{\delta_1}}).
    \label{eq:alpha}
\end{equation}
\textcolor{mycolor}{$\dot{\alpha}$ can be obtained by differentiating Eq. (\ref{eq:alpha}),
\begin{equation}
    \dot{\alpha} = \frac{d}{dt}  \arctan\left(\frac{\dot{\delta}_2}{\dot{\delta}_1}\right)  = \frac{\dot{\delta}_1 \ddot{\delta}_2 - \ddot{\delta}_1 \dot{\delta}_2}{\dot{\delta}_1^2 + \dot{\delta}_2^2}.
    \label{eq:dot_alpha}
\end{equation}
From the kinematic model, we have $\dot{\alpha} = \frac{u}{R} \tan\beta$. With $\dot{\alpha}$ and $u = \sqrt{\dot{\delta}_1^2 + \dot{\delta}_2^2}$, we obtain
\begin{equation}
\beta = \arctan\left( \frac{\dot{\alpha} R}{u} \right) = \arctan\left( \frac{(\dot{\delta}_1 \ddot{\delta}_2 - \dot{\delta}_2 \ddot{\delta}_1) R}{(\dot{\delta}_1^2 + \dot{\delta}_2^2)^{3/2}} \right).
\label{eq:beta}
\end{equation}}
The angular velocity of the steering wheel $\omega$ can be obtained by differentiating Eq. (\ref{eq:beta}),
\begin{equation}
    \begin{aligned}
        \omega = &\dot{\beta}\\
        =& \frac{\begin{aligned}
            &(\dot{\delta_1}\dddot{\delta_2}- \dot{\delta_2}\dddot{\delta_1})(\dot{\delta_1}^2 + \dot{\delta_2}^2)- \\ 
            & 3(\dot{\delta_1}\ddot{\delta_2}-\dot{\delta_2}\ddot{\delta_1})(\dot{\delta_1}\ddot{\delta_1}+\dot{\delta_2}\ddot{\delta_2})
        \end{aligned}
        }{(\dot{\delta_1}^2 + \dot{\delta_2}^2)^3 + (\dot{\delta_1}\ddot{\delta_2}-\dot{\delta_2}\ddot{\delta_1})^2R^2} \times \sqrt{\dot{\delta_1}^2 + \dot{\delta_2}^2} R.
    \end{aligned}
    \label{eq:omega}
\end{equation}
\textcolor{mycolor}{From the above derivations, we observe that given the output $\delta = (x, y)$, which is three times differentiable, we can express other states of the vehicle $[\alpha, \beta]^T$ and the control inputs $[u, \omega]^T$ purely in terms of the $\delta$ and its derivatives. This feature avoids solving differential equations during planning, enabling real-time applications in CAVs \cite{van1998differential}.}

\subsection{Collaborative Decision-Making Formulation}
\label{subsec:collaborative_decision_making_formulation}

Multi-vehicle collaborative decision-making can be modeled as a Decentralized Partially Observable Markov Decision Process (DPOMDP) \cite{bhallaDeepMultiAgent2020} with inter-agent communication (DPOMDP-Com) \cite{oliehoekConciseIntroductionDecentralized2016}.

A DPOMDP is a sequential decision-making problem involving multiple agents (e.g., CAVs) in an environment where information is incomplete and uncertain. Each agent independently takes actions based on its own observations. These actions are taken simultaneously and rely solely on the agent's local observation. The outcomes of these actions provide immediate rewards and generate new observations for each agent, guided by probabilistic models that follow Markov properties and span continuous states, actions, and observation spaces. Mathematically, the DPOMDP model can be described as a tuple \cite{oliehoekConciseIntroductionDecentralized2016}:
\begin{equation}
    \langle \mathbf{A}, \mathbf{S}, \{\mathbf{\Lambda }_i\}, \mathbf{T}, \mathbf{R}, \{\mathbf{O}_i\}, \mathbf{Z}, \mathbf{G} , {\gamma}\rangle.    
\end{equation}
In this model, \( \mathbf{A} \) represents the group of agents and \( \mathbf{S} \) denotes joint state space. It can be decomposed into  environmental state $X^e$ and the joint state of agents $X$, and so we have 
\begin{equation}
    \mathbf{S}=X\times X^e,\ X=\prod_iX^{(i)}.
\end{equation}
Here, $X^{(i)}$ is the state space of agent \( i \), which is a continuous space and environmental state $X^e$ is finite. 

In addition, each agent \( i \) within \( \mathbf{A} \) has its own specific action space \( \mathbf{\Lambda}_i \). Under on-ramp merging scenarios, the agent should make a driving decision at each time step $t$, including changing lane left $\Lambda_\text{left}$, changing lane right $\Lambda_\text{right}$, maintaining the current state $\Lambda_\text{idle}$, and accelerating $\Lambda_\text{acc}$ or decelerating $\Lambda_\text{dec}$. The action space is given by $\mathbf{\Lambda}_i=\{ \Lambda_\text{left}, \Lambda_\text{right}, \Lambda_\text{idle}, \Lambda_\text{acc}, \Lambda_\text{dec}\}$, and the joint action space is 
$\mathbf{\Lambda}=\mathbf{\prod}_{i=1}^n \mathbf{\Lambda}_i$. This means that the action choices of each agent \(i\) form a tuple, containing the independent decisions made by each agent at the same time step \(t\). For example, if there are three agents, one possible joint action could be \( (\Lambda_\text{left}, \Lambda_\text{acc}, \Lambda_\text{dec}) \), indicating that the first agent chooses to change lanes to the left, the second agent chooses to accelerate, and the third agent chooses to decelerate. 

The transition probability function \( \mathbf{T} \) describes the probability of transitioning from one state $s \in \mathbf{S}$ to another state $s^\prime\in\mathbf{S}$ when the set of actions \( \{a_i\} \in \mathbf{\Lambda} \) are taken. 
Moreover, the observation space \( \mathbf{O}_i \) highlights the decentralized aspect of decision-making, and the observation probability function is given by:
\begin{equation}
    \mathbf{Z}: \mathbf{S} \times \prod_{i=1}^n\mathbf{A}_i \rightarrow \Pi\left(\prod_{i=1}^n\mathbf{O}_i\right).
\end{equation}
This function calculates the probability distribution for the observations each agent receives, based on the current state and the actions taken. 

The key to addressing a DPOMDP is to develop a collection of policies \( \pi = \{\pi_1, \pi_2, ..., \pi_n\} \). Each policy $\pi_i$ is a mapping from the agent's observation history and action history into its next action: $a^t_i=\pi_i(H^t_i)$, where $H^t_i=\{(o_i^{1}, a_i^{1}), (o_i^{2}, a_i^{2}), \ldots, (o_i^{t}, a_i^{t})\}$ and $o_i^{t}\in \mathbf{O}_i$ is the observation of agent $i$ at time $t$. In order to find the policies, we can define the value function $\mathbf{V}^{\pi}(s_0)$, which is the expected maximum of rewards $\mathbf{R}$ that an agent can achieve by following a policy $\pi$ from the initial state $s_0$:
\begin{equation}
    \mathbf{V}^{\pi}(s_0)=\mathbf{E}\left[\sum_{t=0}^{\infty}\gamma^t \mathbf{R}(s_t, a_t)|s_0, \pi\right],
\end{equation}
where $\gamma$ is the discount factor, $\gamma=1$ if the problem is finite-horizon, and $0\leq\gamma<1$ if the problem is infinite-horizon.
Then, the optimal policy $\pi^*$ can be obtained by maximizing the value function:
\begin{equation}
    \pi^*=\arg\max_{\pi}\mathbf{V}^{\pi}(s_0).
\end{equation}

Furthermore, the communication among agents can be incorporated into the observations and actions of the agents. If the communication is free, instantaneous and lossless, the DPOMDP model can be simplified to POMDP, which means all agents can get all observations with global information. If the communication has a delay or cost, the agents should decide when to communicate and what to communicate. In this case, we can reformulate the DPOMDP model as a DPOMDP with inter-agent communication (DPOMDP-Com), which contains a set of messages $\mathbf{M}$ \cite{oliehoekConciseIntroductionDecentralized2016}. The complexity of DPOMDP-Com is the same as DPOMDP, but it will be beneficial for decision-making to consider inter-agent communication.

\section{Method}
\label{sec:method}

\begin{figure*}[t]
    \centering
    \includegraphics[width=0.8\linewidth]{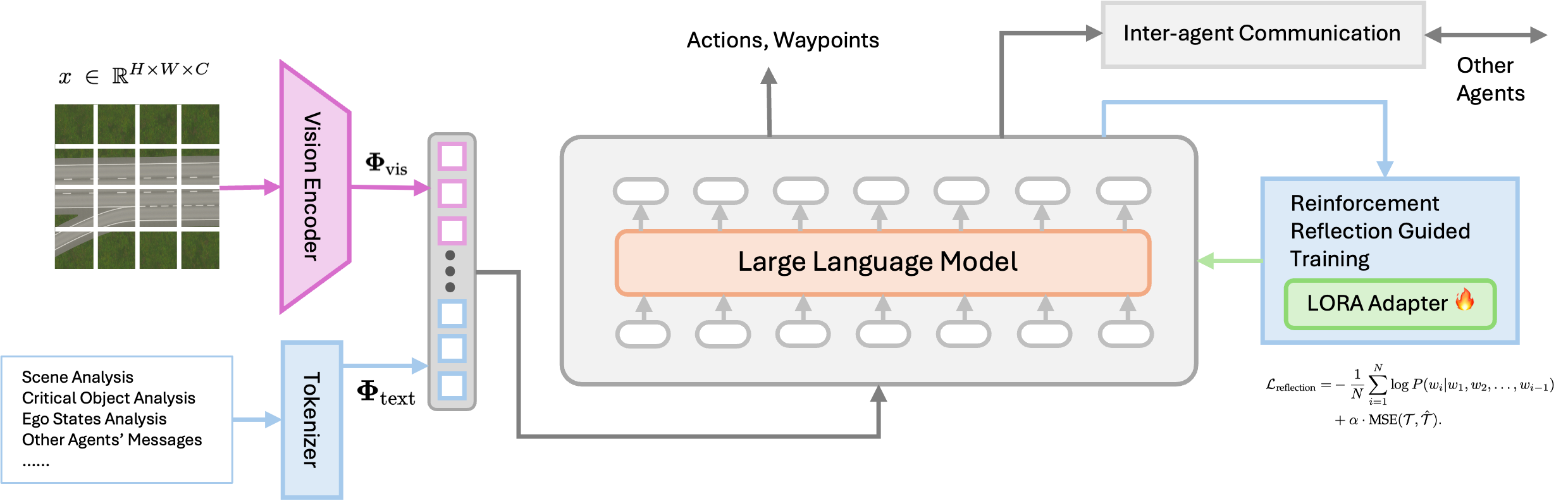}
    \vspace{-3mm}
    \caption{\textbf{Overall Architecture of {AgentsCoMerge}.} The overall architecture of the proposed collaborative driving framework under multi-lane merging scenarios. \textcolor{mycolor}{It includes four parts: 1) scene understanding, 2) planning, 3) inter-agent communication, and 4) reinforcement reflection. First, the agent understands the scene and environments by scene observation and understanding module. Then, the agent plans and makes decisions based on the environments and messages from other agents. Finally, the agent reflects on the decisions and improves its decision-making capability with reinforcement reflection guided training.}}
    \label{fig:overall_architecture}
    \vspace{-5mm}
\end{figure*}

In this section, we propose {AgentsCoMerge}, a novel collaborative driving framework that leverages foundation models to facilitate comprehensive collaborative decision-making for on-ramp merging scenarios in smart transportation systems.

\subsection{Overview}
\label{sec:overall_architecture}

Fig. \ref{fig:overall_architecture} shows our multi-vehicle, closed-loop, and lifelong learning collaborative decision-making framework under multi-lane merging scenarios. It contains four main components: 1) scene observation and understanding (Section \ref{sec:observation}), 2) hierarchical planning (Section \ref{sec:reasoning_engine}), and 3) inter-agent communication (Section \ref{sec:comm}). The scene observation and understanding module is responsible for perceiving the surrounding environments and extracting necessary information. The hierarchical planning module is designed to perform multi-round reasoning based on the inputs and generate the final decision. The inter-agent communication module is responsible for enabling the agents to communicate with each other. In addition, we propose a reinforcement reflection guided training paradigm (Section \ref{sec:reflection}) to enhance the capabilities of reasoning and decision-making, which consists of two steps: 1) planning fine-tuning and 2) reinforcement reflection fine-tuning. The details about each component are discussed in the following sections.

\begin{algorithm}[t]
    \caption{Pseudo-code of AgentsCoMerge Framework}
    \label{alg:agentscomerge}
    \begin{algorithmic}[1]
    \State \textbf{Initialize:} 
    \State $\mathbf{A} = \{a_i\}_{i=1}^N$ \Comment{CAV agent set}
    \State $\mathbf{S} \gets \text{MultiLaneMergingEnv}()$ \Comment{State space}
    
    \For{$e = 1 \ \text{to} \ E_{\text{max}}$}
        \State $\mathbf{o}_0 \gets \mathbf{S}.\text{reset}()$ \Comment{Initial observation}
        \For{$t = 1 \ \text{to} \ T$}
            \State \textbf{Scene Observation \& Understanding:}
                \State $\mathbf{s}_t \gets \mathbf{S}.\text{get\_sensor\_data}(a_i)$
                \State $\mathbf{\Phi}_\text{vis} \gets f_\text{align}(f_\text{encode}(\mathbf{s}_t))$ \Comment{Visual tokens}
                \State $\boldsymbol{\tau}_t \gets g_{\text{text}}(\mathbf{s}_t)$ \Comment{Text tokens}
                \State $\mathbf{r}_{i,t} \gets h_{\text{scene}}(\boldsymbol{\tau}_t, \mathbf{\Phi}_\text{vis})$ \Comment{Scene representation}
            
            \State \textbf{Hierarchical Planning:}
                \State $\mathcal{M}_{i,t} \gets \{\mathbf{m}_{j \rightarrow i}\}_{j \in \mathcal{N}_i}$ \Comment{Received messages}
                \State $\boldsymbol{\phi}_{i,t} \gets h_{\text{LLM}}(\{\mathbf{r}_{i,t}\} \cup \mathcal{M}_{i,t})$ \Comment{Reasoning}
                \State $(\alpha_{i,t}, \mathcal{T}_{i,t}) \gets \psi(\boldsymbol{\phi}_{i,t})$ \Comment{Meta-action \& Traj.}
                \State $\mathbf{x}_{i,t+1} \gets f_{\text{update}}(\alpha_{i,t}, \mathcal{T}_{i,t})$
            
            \State \textbf{Inter-Agent Communication:}
                \If{$a_i \in \mathcal{C}$} \Comment{$\mathcal{C}$: collaborative area}
                    \State $\mathbf{m}_{i,t} \gets h_{\text{LLM}}(\alpha_{i,t})$ \Comment{Generate message}
                    \State $\text{Broadcast}(a_i, \mathbf{m}_{i,t})$
                \EndIf
    
            \State \textbf{Execute \& Observe:}
            \State $\mathbf{S}.\text{execute}(\alpha_{i,t}, \{\mathcal{T}_{i,t}\}_{i=1}^N)$
            \State $[\text{failures}] \gets \mathbf{S}.\text{step}()$ \Comment{Failure cases}
    
            \State \textbf{Reinforcement Reflection:}
                \If{$[\text{failures}]$}
                    \State $\boldsymbol{\rho}_{i,t} \gets h_{\text{LLM}}(\mathcal{T}_{i,t}, c_t)$ \Comment{Reflection}
                    \State $\mathcal{L}_\text{reflection} = \mathcal{L}_{\text{LM}}(\boldsymbol{\rho}_{i,t}, \mathbf{y}_t) + \alpha \|\mathcal{T}_{i,t} - \hat{\mathcal{T}}_{i,t}\|_2^2$
                    \State $\theta \gets \theta - \eta \nabla_{\theta}\mathcal{L}_\text{reflection}$ \Comment{LoRA update}
                \EndIf
        \EndFor
    \EndFor
    \end{algorithmic}
\end{algorithm}

\subsection{Scene Observation and Understanding}
\label{sec:observation}

To facilitate collaboration among agents, it is crucial for CAVs to interpret the environments around them and gather essential data for advanced reasoning processes. To achieve this, an observation system is created for the CAVs, which is designed to capture the context of their surroundings and distill key high-level details such as the number of lanes and the location and velocity of their surrounding vehicles. The collected data from these observations is subsequently passed to the agent's reasoning engine to make decisions. Additionally, this information aids in retrieving relevant memories from the agent's memory module.

\subsubsection{Vision-based Scene Understanding}
In order to enable LLMs to understand visual information and transform it into VLMs. Existing methods generally consist of three parts: 1) an image encoder $f_\text{encode}(\cdot)$ to encode images into latent features, 2) a token alignment module $f_{\text{align}}(\cdot)$ to map visual features to visual tokens that aligned with the space of text tokens, and 3) a large language model $f_\text{LLM}(\cdot)$ to receive visual and text tokens to generate the response. 

\begin{figure}[t]
    \centering
    \includegraphics[width=1\linewidth]{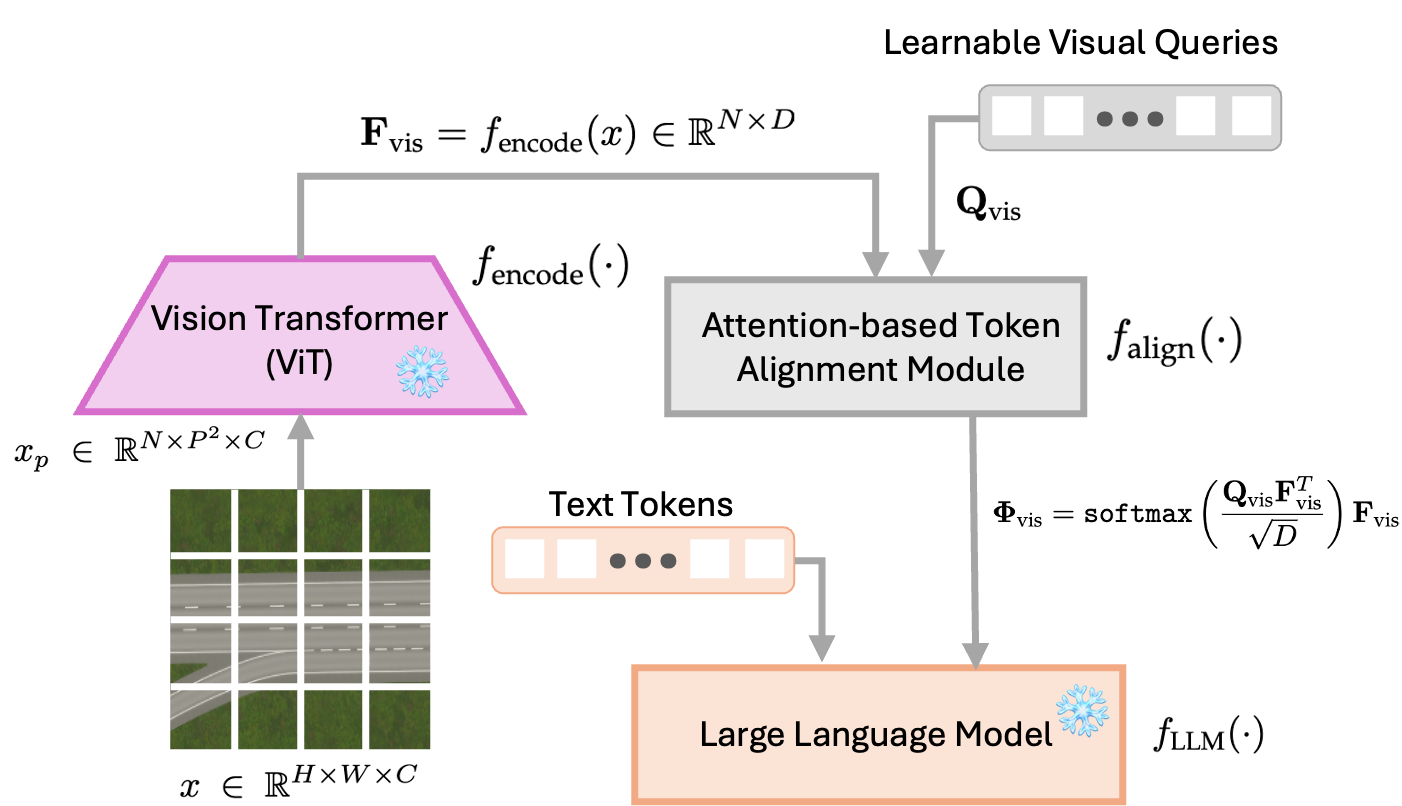}
    \vspace{-5mm}
    \caption{Design of the visual-based scene understanding module.}
    \label{fig:visual-encoder}
    \vspace{-5mm}
\end{figure}

In our design, as shown in Fig. \ref{fig:visual-encoder}, we leverage the Vision Transformer (ViT) \cite{dosovitskiyImageWorth16x162021} as the image encoder $f_\text{encode}(\cdot)$ to encode images as visual features. Specifically, an image $x\in \mathbb{R}^{H\times W\times C}$ is first resized into a sequence of flattened 2D patches $x_p \in \mathbb{R}^{N\times P^2 \times C}$. $(H,W)$ is the spatial resolution of the image, $C$ is the number of channels, $(P,P)$ is the resolution of each patch, and $N$ is the number of patches, $N=(H\times W)/{P^2}$, which also serves as the input sequence length for the Transformer. The patches are then linearly embedded into a lower-dimensional representation, which stands for the visual features $\mathbf{F}_\text{vis}$.
\begin{equation}
    \mathbf{F}_\text{vis} = f_\text{encode}(x) \in \mathbb{R}^{N\times D},
\end{equation}
where $D$ is the embedding dimension. The positional encoding is added to the embedded patches to preserve the spatial information. The Transformer is then applied to the embedded patches to capture the long-range dependencies between the patches. 

After obtaining the visual features $\mathbf{F}_\text{vis}$, we then feed them into the token alignment module $f_{\text{align}}(\cdot)$ to align the visual tokens to the space of text tokens. Specifically, we design a single-layer cross-attention module that is initialized with random values. It employs a set of trainable vectors (Embeddings) $\mathbf{Q_\text{vis}}$ as query vectors, while using the visual features $\mathbf{F}_\text{vis}$ from the visual encoder as keys and the values in the cross-attention operations. Then we can obtain the cross attention matrix $\mathbf{\Phi }_\text{vis}$ \cite{vaswaniAttentionAllYou2017}:
\begin{equation}
    \mathbf{\Phi}_\text{vis} = \mathtt{softmax}\left(\frac{\mathbf{Q}_\text{vis}\mathbf{F}_\text{vis}^T}{\sqrt{D}}\right)\mathbf{F}_\text{vis}.
\end{equation}
Then, the cross attention matrix $\mathbf{\Phi}_\text{vis}$ is used as visual tokens and concatenated with the text tokens to feed into the LLM to generate the response. In our implementation, we have made only the single-layer cross-attention module trainable, while all other parameters are kept frozen to entirely preserve the knowledge from the pre-trained models.

\subsubsection{Text-based Scene Understanding}

In text-based scene understanding, since LLM can directly understand the text, we can shift our focus away from architectural design (in vision-based understanding) to prompt engineering. By feeding well-designed text-based information into LLM, we can efficiently generate responses. The text-based information includes the high-level scene analysis, the low-level critical object analysis, and the structured description of the environment.

\textbf{Scene analysis.}
The scenes are very important for driving, and different scenes may have different driving decisions. For example, the driving behaviors in urban areas are different from those in rural areas. Specifically, in multi-lane merging scenarios, we consider the road structure, the number of lanes, and a CAV's current lane. In addition, the ego state of the CAV is also one of the basis of decision-making, such as the current speed and the related position on the road.Fig. \ref{fig:qualitative-analysis} shows an example of the scene analysis.

\textbf{Critical object analysis.} Apart from the high-level scene analysis, it is also crucial to identify the critical objects in the scene, such as the surrounding vehicles and pedestrians. In contrast to conventional autonomous driving perception systems that detect all objects, our approach is to specifically recognize key objects that have the highest potential to affect the current situation around merging lanes, which is like the human's cognitive process during driving. 

\textbf{Information summary and encoding.} After obtaining the high-level scene analysis and low-level critical object analysis, a CAV agent will summarize the information into a comprehensive structured description, which will be fed into the reasoning engine for further processing.

\vspace{-3mm}
\subsection{Hierarchical Planning}
\label{sec:reasoning_engine}

Reasoning is an essential and core human capability, crucial for making both everyday and intricate decisions. Traditional data-driven approaches rely directly on sensory data, such as object detection and semantic segmentation results, to determine driving actions (such as turning left or right, or adjusting speed), which often lacks transparency and struggles with complex situations and rare events. Drawing inspiration from human reasoning capabilities, we propose a reasoning engine for a CAV agent.

\textbf{Reasoning Process.}  In order to obtain decisions, an agent needs to reason over all the information. Given the complexity and diversity of the information, a robust reasoning engine is crucial for enabling the agent to make effective and accurate decisions. To enhance reasoning capabilities, \textcolor{mycolor}{we directly employ LLM as the reasoning engine with carefully designed prompts combined with the observation tokens (visual tokens and text tokens) to process all the information and generate the decision.}

Typically, human drivers break down their decision-making process into a series of sequential steps based on a coherent order. This includes recognizing and pinpointing important objects, anticipating their potential movements and interactions, and formulating a plan for their own actions based on the gathered information. 
Inspired by this, we employ the innovative chain-of-thought prompting technique introduced in \cite{weiChainofThoughtPromptingElicits}. This method enables an LLM to break down complex problems into a sequence of sub-problems, solving them step by step by producing a series of intermediate outputs before arriving at the final decisions, so it can offer an interpretable view into the model's behavior, indicating how it may have reached a specific conclusion and providing opportunities to identify where the reasoning process may have faltered.
We adopt this strategy due to the inherent complexity and variability of collaborative driving scenarios, which could lead to erroneous or hallucinatory outcomes if the LLM generates decision results directly.

\textbf{Meta-action Decision.} Upon completing the reasoning process, the agent generates the final decision, which is subsequently decoded into specific meta-actions for execution by the agent.
In addition, to determine the future actions of the ego CAV over a specified time period, we construct a series of meta-actions. Each meta-action in this sequence plays a crucial role, collectively shaping the strategic movement of the CAV within the environment.

\textbf{Trajectory Planning.} After obtaining the meta-action decision, the agent needs to refine the decision into a specific trajectory $\mathcal{T}=\{(x_1, y_1), (x_2, y_2), \ldots, (x_t, y_t)\}$. Note that we can directly represent the trajectory as a sequence of words \cite{maoGPTDriverLearningDrive2023}:
\begin{equation}
    \begin{aligned}
        \mathcal{T}= &\mathtt{Tokenizer}(\{(x_1, y_1), (x_2, y_2), \ldots, (x_t, y_t)\})\\
        = & \{w_1, w_2, \ldots, w_t\}, 
    \end{aligned}
\end{equation}
where $\text{Tokenizer}(\cdot)$ is a function that converts the trajectory into a sequence of words and $w_i$ is the $i$-th word in the sequence. With this language representation, we can reformulate the trajectory planning problem as a language generation problem \cite{maoGPTDriverLearningDrive2023}:
\begin{equation}
    \mathcal{L}_{\text{LM}}=-\sum^N_{i=1}\log P(\hat{w}_i|w_1, w_2, \ldots, w_{i-1}),
    \label{eq:lm_loss}
\end{equation}
\textcolor{mycolor}{where \( w \) and \( \hat{w} \) represent the words from the planned trajectory \( \mathcal{T} \) and the input tokens (history trajectory, scene understanding, etc.), respectively. By focusing on increasing the probability \( P \) of the words \( \hat{w} \) derivable from the input tokens, the LLM can create driving trajectories more precisely. In addition, we can reformulate the language generation problem as an optimization problem:}
\begin{equation}
    \min_\theta \mathcal{L}_{\text{LM}}(\theta)=-\sum^N_{i=1}\log P_\theta(\hat{w}_i|w_1, w_2, \ldots, w_{i-1}).
    \label{eq:lm_loss}
\end{equation}
In order to optimize this problem, we can use gradient descent to update the parameters \( \theta \) of the language model \( P_\theta \) to minimize the loss function \( \mathcal{L}_{\text{LM}} \).    
\begin{figure}[t]
    \centering
    \includegraphics[width=1\linewidth]{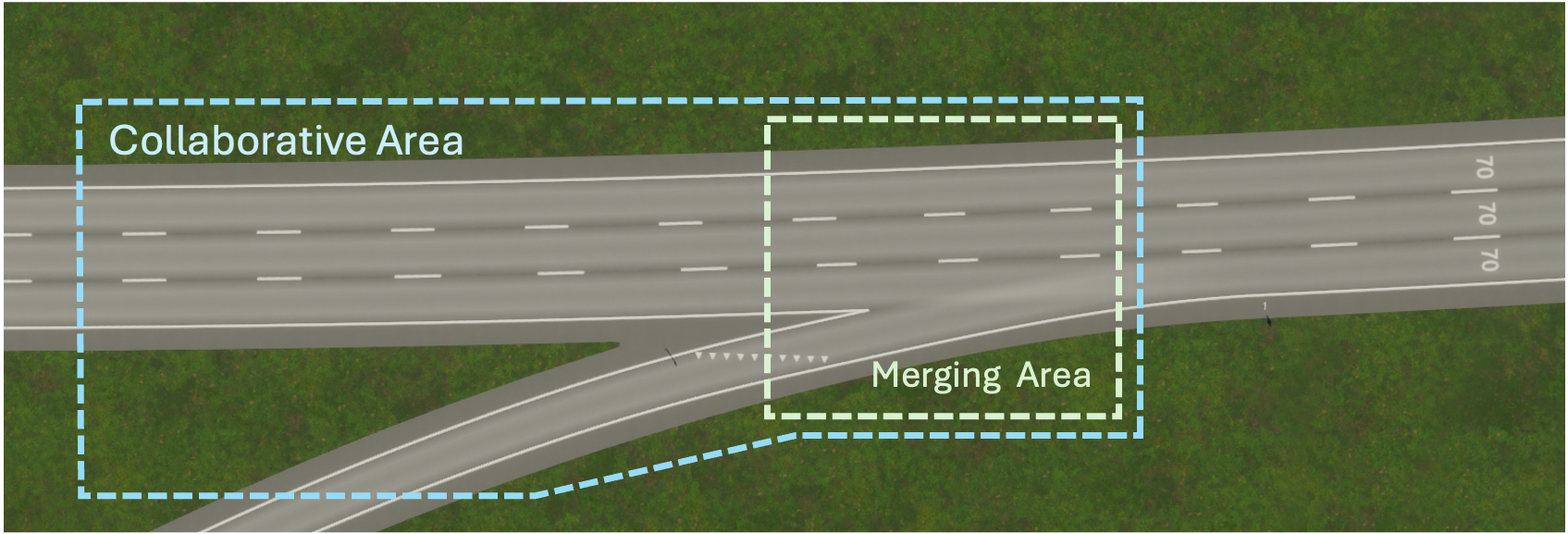}
    \vspace{-5mm}
    \caption{Illustration of the collaborative area and the merging area in multi-lane merging scenario.}
    \label{fig:collaborative_merging_area}
    \vspace{-5mm}
\end{figure}

\textcolor{mycolor}{\textbf{Output of hierarchical planning.} The output of the hierarchical planning is the meta-action decision and the trajectory, as well as the reasoning process.}

\subsection{Inter-agent Communication}
\label{sec:comm}

In order to realize a multi-agent system, it is crucial for agents to communicate with each other, and there are listener agents and speaker agents. The communication between different agents can be classified into different types \cite{bhallaDeepMultiAgent2020}.
\begin{enumerate}
    \item \textit{Cheap Talk} refers to a form of communication model where messages are sent without any cost and are not necessarily binding or influential to the receiver's actions. It is characterized by the freedom for agents to ignore the received messages, and is referred to as ``doing by talking" because it allows for potentially inconsequential communication.
    \item \textit{Signalling Games} differ in that communication occurs through actions rather than explicit messages. Agents signal their intentions or states through their behaviors, which indirectly convey information. This form of communication is described as ``talking by doing," where the actions themselves are the messages.
    \item \textit{Speech Act Theory} is a structured form of communication similar to Cheap Talk, yet it is designed to transmit actionable information that can significantly influence the listener's policy. This model is particularly suitable for cooperative environments, where the communication is expected to be transparent (self-revealing) and consistent with the speaker's own actions (self-committing). The theory categorizes actions into two types: locutionary (behavioral) actions that alter the environment, and illocutionary (message) actions that do not directly change the environment but instead aim to influence the listener's behavior. The primary objectives of this communication strategy are to improve speaker consistency and enable instantaneous coordination, ensuring that the messages not only mirror the speaker's actions but also effectively guide the listener in achieving collaborative objectives.
\end{enumerate}

In this paper, we mainly focus on the Speech Act Theory as the communication model. In addition, as the discussion of DPOMDP-Com in Section \ref{subsec:collaborative_decision_making_formulation}, if the communication has a delay of cost, agents should decide \textit{when} to communicate and \textit{what} to communicate. Firstly, in the multi-lane merging scenario, we set a collaborative area at the entrance of a ramp, where agents can communicate with each other once a CAV enters this area. The collaborative area is shown in Fig. \ref{fig:collaborative_merging_area}. As for the content of communication, we incorporate the generation of messages into the reasoning process, seamlessly generating self-revealing and self-committing messages. In this case, the issues of \textit{when} to communicate and \textit{what} to communicate are naturally addressed.

\begin{figure}[t]
    \centering
    \includegraphics[width=1\linewidth]{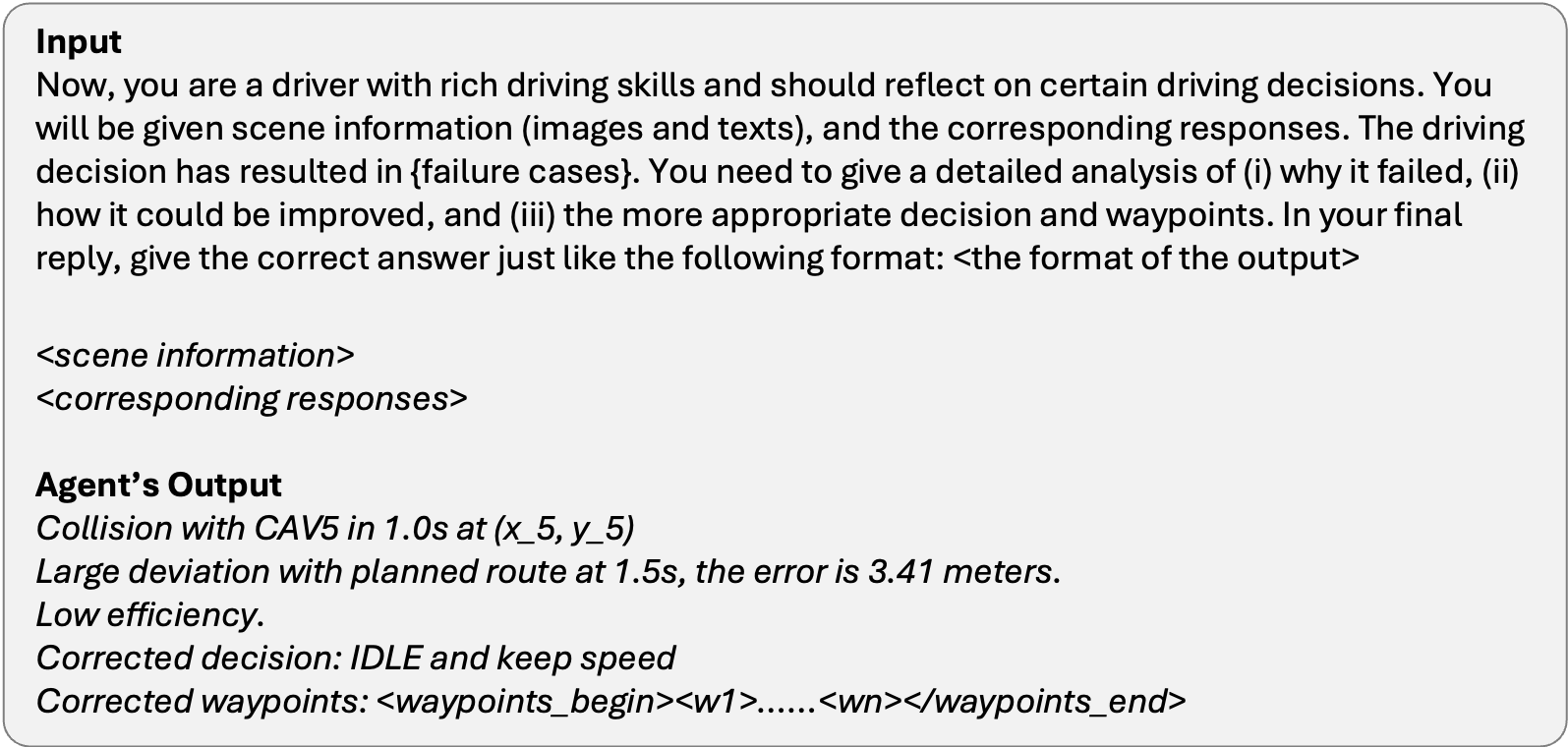}
    \vspace{-5mm}
    \caption{Example of the reinforcement reflection reasoning process.}
    \label{fig:reinforcement_example}
    \vspace{-5mm}
\end{figure}

\subsection{Reinforcement Reflection Guided Training}
\label{sec:reflection}

In order to become an expert in a specific area, an individual needs to learn from previous experiences. This involves the capability to reflect on past errors and thoroughly understand the causes behind these mistakes. In so doing, the individual can ``train" his or her brain and reduce the likelihood of making the same mistakes in the future. For example, a human driver who has experienced a near-accident while merging onto a highway will likely be more cautious in similar situations in the future. Because this incident makes the driver feel nervous or afraid, which deepens the impression of the incident. Inspired by this, we just then consider whether we can also deepen the impression of the dangerous or wrong decisions of an agent just like what a human driver does, thereby improving the agent's decision-making ability.

Aligning with this motivation, we propose reinforcement reflection guided training. This training paradigm consists of two steps: 1) planning fine-tuning and 2) reinforcement reflection fine-tuning. Specifically, in the first step, we fine-tune the model by Low-Rank Adaptar (LoRA) \cite{huLORALOWRANKADAPTATION2022}, which freezes the model weights and introduces trainable matrices based on rank decomposition into each layer of the Transformer. During fine-tuning, each data sample is formatted to include an input prompt and a corresponding answer. The parameters of the adapter are optimized by the loss function provided by Eq. (\ref{eq:lm_loss}).

In the second step, we utilize detailed textual reflection about errors in meta-actions and waypoint prediction to enhance the learning process of the agent. This approach allows the agent to gain an impression from successes and failures, which can offer a particularly instructive form of supervision. 
Specifically, we first define the failure cases including \textit{collisions, deviation of route, low efficiency}, and \textit{low comfort}, if these failure cases occur, we will prompt the agent to reflect on its decisions and predictions. The example of this process is shown in Fig. \ref{fig:reinforcement_example}. \textcolor{mycolor}{The failure cases are described as follows:}
\begin{itemize}
    \setlength{\itemsep}{0pt}
    \setlength{\parskip}{0pt}
    \setlength{\parsep}{0pt}
    \item \textcolor{mycolor}{\textit{Collision}: To detect potential collisions, we use the kinematic bicycle model ${K}_{veh}$ to predict the future states $\mathbf{S}^i_{t+1}=(x_{t+1}^i, y_{t+1}^i, \alpha_{t+1}^i, \beta_{t+1}^i)$ of surrounding vehicles based on their current states $\mathbf{S}^i_t=(x_t^i, y_t^i, \alpha_t^i, \beta_t^i)$ and control inputs $\mathbf{C}_t^i = [u_t^i, w_t^i]^T$. We then compute the intersections over union (IoU) between the ego vehicle's bounding box $B_t^s$ and other vehicles' bounding boxes $O_t^i$ at each predicted waypoint. A collision is detected if $\text{IoU}(B_t^s, O_t^i) > \varepsilon_{col}$, where $\varepsilon_{col}$ is a predefined collision threshold.}
    \item \textcolor{mycolor}{\textit{Route Deviation}: We further compute the deviation of the predicted waypoints $\mathcal{T}_t^s$ from $N_c$ points along the centerline of the planned route in order to force the ego vehicle stay on road. A route deviation is detected if $\text{Dev\_Plan}_t = \mathbf{1}(\min_{i\in N_c}(\|\mathcal{T}_t^s - \mathcal{T}_t^p\|_2) > \varepsilon_P)$, where $\mathcal{T}_t^p$ is the $i$-th point along the center line. We examine deviations by verifying whether the distance between the predicted waypoint and its nearest point along the centerline of the planned route is larger than the threshold $\varepsilon_P$.}
    \item \textcolor{mycolor}{\textit{Low Efficiency}: We evaluate the driving efficiency using a score Efficiency Score (ES) that compares the ego vehicle's velocity $v_{ego}$ with a reference velocity $v_0$, which is either the average velocity of surrounding vehicles $v_{avg}$ or the speed limit $v_{lmt}$. The efficiency score is calculated as follows
    \begin{equation}
    \text{ES} = \begin{cases} 
    1.0, & \text{if } v_{ego} \geq v_0 \\
    v_{ego}/v_0, & \text{else}
    \end{cases}
    \label{eq:efficiency_score}
    \end{equation}
    where $v_0 = \min\{v_{avg}, v_{lmt}\}+\sigma$. $\sigma$ is the standard deviation of $v_0$, we use $\sigma = 0$ in this study and can choose $2\sigma$ or $3\sigma$ according to the past observations. A low efficiency state is detected when $\text{ES} < \varepsilon_E$ for the threshold $\varepsilon_E$.}
    \item \textcolor{mycolor}{\textit{Low Comfort}: The comfort of driving behavior is measured using a Comfort Score (CS) that considers both lateral and longitudinal aspects of motion:
    \begin{equation}
    \text{CS} = (s_{x_a} + s_{y_a} + s_{x_j} + s_{y_j})/4
    \label{eq:comfort_score}
    \end{equation}
    where $s_{x_a}$ and $s_{y_a}$ are the reference values of lateral and longitudinal acceleration, respectively, and $s_{x_j}$ and $s_{y_j}$ are the reference values of lateral and longitudinal jerk. A low comfort state is triggered when $\text{CS} < \varepsilon_C$ for the threshold $\varepsilon_C$.}
\end{itemize}
The agent is then trained to minimize the loss function, defined as the weighted sum over the language cross-entropy loss and the waypoint prediction mean square error (MSE) loss:
\begin{equation}
    \begin{aligned}
        \mathcal{L}_\text{reflection} = & -\frac{1}{N}\sum_{i=1}^{N}\log P(\hat{w}_i|w_1, w_2, \ldots, w_{i-1}) \\
            & + \alpha\cdot \text{MSE}(\mathcal{T}, \hat{\mathcal{T}}).
    \end{aligned}
\end{equation}
To optimize this problem, we should minimize the loss function:
\begin{equation}
\min_\theta \mathcal{L}_\text{reflection}(\theta).
\end{equation} 
At each iteration, we first compute loss $\mathcal{L}_\text{reflection}$ by propagating the input through the model, then we compute the gradients of the loss with respect to the model parameters $\theta$. Finally, we update the model parameters $\theta$ using gradient descent until convergence:
\begin{equation}
    \theta \leftarrow \theta - \eta \nabla_\theta \mathcal{L}_\text{reflection}(\theta).
\end{equation}


\begin{table*}[t]
    \centering
    
    \caption{\textbf{Open-loop evaluation in the nuScenes dataset}. The results are reported in terms of L2 error and collision rate. }
    \vspace{-3mm}
    \renewcommand\arraystretch{1.2}
    \resizebox{1\textwidth}{!}{
    \setlength{\tabcolsep}{5mm} 
    \begin{tabular}{p{4cm}|cccc|cccc}
        \toprule
        \multirow{2}{*}{Method} & \multicolumn{4}{c|}{L2 (m) $\downarrow$} & \multicolumn{4}{c}{Collision (\%) $\downarrow$} \\
         & 1s & 2s & 3s & Avg. & 1s & 2s & 3s & Avg. \\
        \midrule
        NMP \cite{zengEndToEndInterpretableNeural2019} & - & - & 2.31 & - & - & - & 1.92 & -  \\
        SA-NMP \cite{zengEndToEndInterpretableNeural2019} & - & - & 2.05 & - & - & - & 1.59 & -  \\
        FF \cite{huSafeLocalMotion2021} & 0.55 & 1.20 & 2.54 & 1.43 & 0.06 & 0.17 & 1.07 & 0.43 \\
        EO \cite{khuranaDifferentiableRaycastingSelfsupervised2022} & 0.67 & 1.36 & 2.78 & 1.60 & 0.04 & 0.09 & 0.88 & 0.33 \\
        ST-P3 \cite{huSTP3EndtoEndVisionBased2022} & 1.33 & 2.11 & 2.90 & 2.11 & 0.23 & 0.62 & 1.27 & 0.71 \\
        UniAD \cite{huPlanningorientedAutonomousDriving2023} & 0.48 & 0.96 & 1.65 & 1.03 & 0.05 & 0.17 & \textbf{0.71} & 0.31 \\
        GPT-Driver \cite{maoGPTDriverLearningDrive2023}& 0.27 & 0.74 & 1.52 & 0.84 & 0.77 & 0.15 & 1.00 & 0.44 \\ 
        \midrule
        AgentsCoMerge (Ours) &\textbf{0.25}&\textbf{0.54}&\textbf{0.99}&\textbf{0.59}&\textbf{0.00}&\textbf{0.12}&0.75&\textbf{0.29} \\
        \bottomrule
        \end{tabular}
    }\label{tab:open-loop}
    \vspace{-5mm}
\end{table*}

\begin{table}[t]
    \centering
    
    \caption{\textbf{Close loop evaluation on the LimSim++ Benchmarks.} The results are reported in terms of Comfort Score (CS), Efficiency Score (ES), Safety Score (SS), and Driving Score (DS).}
    \vspace{-3mm}
    \label{tab:closed-loop}
    \renewcommand\arraystretch{1.2}
    \resizebox{1\linewidth}{!}{
    \begin{tabular}{p{3cm}|cccc}
    \toprule
    Method &  CS $\uparrow$ &ES $\uparrow$& SS $\uparrow$& DS $\uparrow$\\
    \midrule
    GPT-4 &0.802&0.456&0.928&77.877 \\
    GPT-4 Turbo &0.797&0.424&0.907&71.697 \\
    GPT-4o &0.804&0.420&0.945&71.507 \\
    Claude-3-Sonnet &0.778&0.424&0.907&71.697 \\
    Claude-3-Opus&0.834&0.246&\textbf{1.000}&74.257 \\
    GLM-3 Turbo &0.807&0.484&0.877&66.591 \\
    GLM-4 &0.787&0.417&0.938&70.679 \\
    Gemini Pro &\textbf{0.840}&0.312&0.959&75.825 \\
    Gemini-1.5 Pro &0.814&0.248&0.976&72.857 \\
    \midrule
    AgentsCoMerge (Ours) &0.828&\textbf{0.499}&0.925&\textbf{79.425} \\
    \bottomrule
    \hline
    \end{tabular}
    }
    \vspace{-3mm}
\end{table}

\vspace{-5mm}
\section{Experiments}
\label{sec:experiments}

In this section, we conduct experiments to evaluate the performance of our proposed method. We first introduce the datasets and evaluation metrics used in our experiments (Section \ref{sec:benchmarks}). Then, we compare our method with the state-of-the-art methods in terms of both quantitative and qualitative results (Section \ref{sec:comparison}). Finally, we conduct ablation studies to investigate the effectiveness of each component in our proposed method (Section \ref{sec:ablation}).

\subsection{Benchmarks and Metrics}
\label{sec:benchmarks}


In order to demonstrate the performance of the proposed method, we evaluate the performance of our method in both closed-loop and open-loop.

\textbf{Closed-loop Benchmarks.} In the closed-loop evaluation, we leverage the LimSim++ simulator \cite{fuLimSimClosedLoopPlatform2024}. LimSim++ is a platform that can simulate the fine-grained dynamic environment and can provide a continuous simulation under different traffic scenarios. In addition, it can allow agents to interact with different vehicles. 

In order to evaluate the performance of our method in this simulator, we leverage the following metrics: Comfort Score (CS), Efficiency Score (ES), Safety Score (SS), and Driving Score (DS) \cite{fuLimSimClosedLoopPlatform2024}. CS measures the comfort of the driving behavior and can be formulated as Eq. \ref{eq:comfort_score}.
ES measures the efficiency of the driving behavior and can be formulated as Eq. \ref{eq:efficiency_score}.
SS measures the safety of the driving behavior assessed by time to conflict (TTC), which estimates the time remaining before a potential collision occurs between two agents (e.g., vehicles, pedestrians) if both maintain their current velocity and trajectory. TTC is calculated by ${d}/{v_{\text{rel}}}$, where $d$ is the current distance between the ego vehicle and the object ahead, $v_{\text{rel}} = v_{\text{ego}} - v_{\text{lead}}$ is relative speed. If $v_{\text{rel}} \leq 0$, TTC is considered infinite (no collision under current conditions). Therefore, $\text{TTC} \in [0, \infty)$.

If TTC is less than a specific threshold, it means that the agent is in a dangerous situation. SS can be formulated as
\begin{equation}
    \text{SS}=
    \left\{\begin{aligned}
        &1.0, & \text{if}\ t_{\text{ego}} \geq t_{t}\\
        &t_\text{ego}/t_t,&\text{else}
    \end{aligned}\right.,
    \label{eq:safety_score}
\end{equation}
where $t_{\text{ego}} \in [0, \infty)$ is the TTC of an agent and $t_t$ is the threshold. Finally, DS measures the overall driving behavior and can be formulated as
\begin{equation}
    \text{DS} = \alpha^{\lambda_1} \cdot \beta^{\lambda_2} \cdot(k_1 \cdot \text{CS} + k_2 \cdot \text{ES} + k_3 \cdot \text{SS}),
\end{equation}
where $\alpha \in [0, 1)$ and $\beta \in [0, 1)$ are the penalty factors for collisions and speed violations (speed violation is triggered when the speed exceeds 11.11m/s), respectively. $\lambda_1$ and $\lambda_2$ represent the times at which two different types of punishable behaviors occur throughout the trajectory. $k_1$, $k_2$, and $k_3$ are the weights of CS, ES, and SS, respectively.

\begin{table}[t]
    \centering
    \caption{\textbf{Ablation study of the effect of communication.} `w/o' means without communication, and `w/' means with communication. 
    The results are reported in terms of Comfort Score (CS), Efficiency Score (ES), Safety Score (SS), and Driving Score (DS).}
    \vspace{-3mm}
    \setlength{\tabcolsep}{2.5mm} 
    \renewcommand\arraystretch{1.2}
    \resizebox{1\linewidth}{!}{
    \begin{tabular}{p{2.5cm}|cccc}
    \toprule
    Metrics &  CS $\uparrow$ &ES $\uparrow$& SS $\uparrow$& DS $\uparrow$\\
    \midrule
    w/o &0.783&0.403&0.837&71.500 \\
    w/ &0.828&{0.499}&0.925&{79.425} \\ 
    \midrule
    Improvement (\%)  $\uparrow$ &5.75&23.82&10.51&11.08 \\ 
    \bottomrule
    \end{tabular}
    }
    \label{tab:ablation-communication}
    \vspace{-3mm}
\end{table}

\begin{table}[t]
    \centering
    
    \caption{\textbf{Ablation study of the effect of reinforcement reflection guided training.} `w/o' means without the reinforcement reflection guided training, and `w/' means with the reinforcement reflection guided training. The results are reported in terms of L2 error and collision rate.}
    \vspace{-3mm}
    \renewcommand\arraystretch{1.2}
    \resizebox{1\linewidth}{!}{
    \setlength{\tabcolsep}{4mm} 
    \begin{tabular}{p{2cm}|cccc}
        \toprule
        \multirow{2}{*}{Metrics} & \multicolumn{4}{c}{L2 (m) $\downarrow$}  \\
         & 1s & 2s & 3s & Avg.  \\
        \midrule
        w/o &{5.11}&{6.45}&{8.08}&{6.55} \\
        w/ &{0.25}&{0.54}&{0.99}&{0.59} \\
        \midrule
        \midrule
        \multirow{2}{*}{Metrics} & \multicolumn{4}{c}{Collision (\%)$\downarrow$}  \\
         & 1s & 2s & 3s & Avg.  \\
        \midrule
        w/o &{11.17}&{13.00}&{13.19}&{12.45} \\
        w/ &{0.00}&{0.12}&{0.75}&{0.29} \\
        \bottomrule
        \end{tabular}
    }\label{tab:ablation-rrgt}
    \vspace{-3mm}
\end{table}

\textbf{Open-loop Benchmarks.} In the open-loop evaluation, we leverage the nuScenes \cite{caesarNuScenesMultimodalDataset2020} dataset. The nuScenes dataset is a large-scale dataset for autonomous driving, which contains 1000 scenes, each of which contains 20 seconds of data. In order to evaluate our method in this dataset, we leverage L2 error and the collision rate. The L2 error measures the distance between the predicted trajectory and the ground truth trajectory, which is formulated as follows \cite{mao2024a}.
\begin{equation}
    \text{L2} = \frac{1}{N\times K}\sum_{j=1}^N\sum_{i=1}^{K} \sqrt{(x_i^j - \hat{x}_i^j)^2 + (y_i^j - \hat{y}_i^j)^2},
\end{equation}
where $N$ is the number of scenarios, $K$ is the number of waypoints in each scenario, $(\hat{x}_i^j, \hat{y}_i^j)$ is the $i$-th human driving waypoints of the $j$-th scenarios, and $({x}_i^j, {y}_i^j)$ is the $i$-th predicted waypoints of the $j$-th scenario.

The collision rate can be expressed as the ratio of the number of times a planned trajectory collides with other objects in a scene to the duration of the planned trajectory. The collision rate is formulated as follows \cite{mao2024a}.
\begin{equation}
    \text{ColliRate} = \frac{1}{N}\sum_{j=1}^N\frac{C^j}{T},
\end{equation}
where $N$ is the number of scenarios, $C^j$ is the number of collisions in the $j$-th scenario, and $T$ is the duration of the planned trajectory.

In addition, we train and evaluate the performance on HighD dataset \cite{8569552}. The HighD dataset is a naturalistic vehicle trajectory dataset recorded from drone videos on German highways. It consists of 60 video recordings (instances), covering a cumulative driving duration of approximately 16.5 hours. The dataset involves over 110,500 vehicles, totaling around 45,000 km of driving data, and it captures approximately 5,600 lane-change maneuvers. The dataset was collected at six distinct highway sections (stretches), enabling evaluation across various road conditions. In order to learn the HighD dataset, we split the dataset into training set (the first 50 recordings) and testing set (the last 10 recordings). Then, we extract the trajectories into 10s and split them into 6s of past and 4s of the future trajectories pairs, which are used to train or evaluate the model. The 6s of past trajectories are feed into the model to predict the 4s of the future trajectories. In order to calculate the loss, we did not use the traditional RMSE loss function, because in our design, the trajectories are generated by the LLM, and we reformulate the trajectory planning problem as a language generation problem as stated in Section 4.3. The objective function is formulated as Eq. \ref{eq:lm_loss}. In order to evaluate the performance on the HighD dataset, we use the root mean square error (RMSE) as evaluation metrics. The RMSE can measure the average deviation between the predicted and the actual values. It is calculated as:
\begin{equation}
    \text{RMSE} = \sqrt{\frac{1}{N} \sum_{i=1}^{N} (y_i - \hat{y}_i)^2}
\end{equation}
where $N$ is the number of samples, $y_i$ is the actual value, and $\hat{y}_i$ is the predicted value.

\begin{table}[t]
    \centering

    \caption{Performance Evaluation on HighD Dataset}
    \vspace{-3mm}
    \label{tab:rmse}
    \resizebox{\linewidth}{!}{
    \begin{tabular}{c|cccc|c}
        \toprule
        \multirow{2}{*}{Method} & \multicolumn{4}{c|}{RMSE in Prediction Horizons (s)} & \multirow{2}{*}{Avg.} \\
        & 1s & 2s & 3s & 4s & \\
        \midrule
        CS-LSTM \cite{8575356} & 0.61 & 1.27 & 2.09 & 3.10 & 1.77 \\
        Dual-Trans \cite{10061336} & 0.41 & 0.79 & 1.11 & 1.40 & 0.93 \\
        WSiP \cite{Wang_Wang_Yan_Wang_2023} & 0.20 & 0.60 & 1.21 & 2.07 & 1.02 \\
        Social GAN \cite{8578338} & 0.30 & 0.78 & 1.46 & 2.34 & 1.22 \\
        \midrule
        Ours & 0.25 & \textbf{0.53} & \textbf{0.91} & \textbf{1.22} & \textbf{0.73} \\
        \bottomrule

    \end{tabular}
    }
    \vspace{-3mm}
\end{table}

\subsection{Implementation Details}
\label{sec:implementation}

We leverage Qwen-7B \cite{bai2023qwentechnicalreport} as our pre-trained language model, which contains 7.7 billion parameters. 
For closed-loop evaluation, we set the weights of CS, ES, and SS as $k_1=0.25$, $k_2=0.25$, and $k_3=0.5$, respectively. The penalty factors for collisions and speed violations are set on $\alpha=0.6$ and $\beta=0.9$, respectively. In addition, we set the TTC threshold in Eq. \ref{eq:safety_score} as $t_t=5s$.

\subsection{Comparison Analysis}
\label{sec:comparison}

In this section, we conduct comparison experiments to evaluate the performance of our proposed method against the previous state-of-the-art methods.

\textbf{Closed-loop Evaluation.} In order to evaluate the decision-making performance of our proposed method {AgentsCoMerge}, we leverage SUMO to build a multi-lane merging scenario and evaluate it in the LimSim++ simulator. This scenario contains a main road with 3 lanes and a ramp with a single lane. We need to control the ego vehicle to merge into the main road from the ramp and control the other vehicles to drive and interact with the ego vehicle. We compare our method with other methods, including GPT-4, GPT-4 Turbo, GPT-4o, Claude-3-Sonnet, Claude-3-Opus, GLM-3 Turbo, GLM-4, Gemini Pro, and Gemini-1.5 Pro, and we report the results in terms of CS, ES, SS, and DS.
The results are shown in Table \ref{tab:closed-loop}. As we can see,  Claude-3-Opus excels in SS with a perfect score of 1.000 but has a lower ES of 0.246. Gemini Pro and Gemini-1.5 Pro perform well in SS with scores of 0.959 and 0.976, respectively, and Gemini Pro performs the best in CS with a score of 0.840, but they have lower ES.
However, our proposed method {AgentsCoMerge} offers a competitive profile with the highest DS of 79.425 and a strong CS of 0.828, showing it as a top contender in terms of driving performance.

\begin{figure}[t]
    \centering
    \includegraphics[width=.9\linewidth]{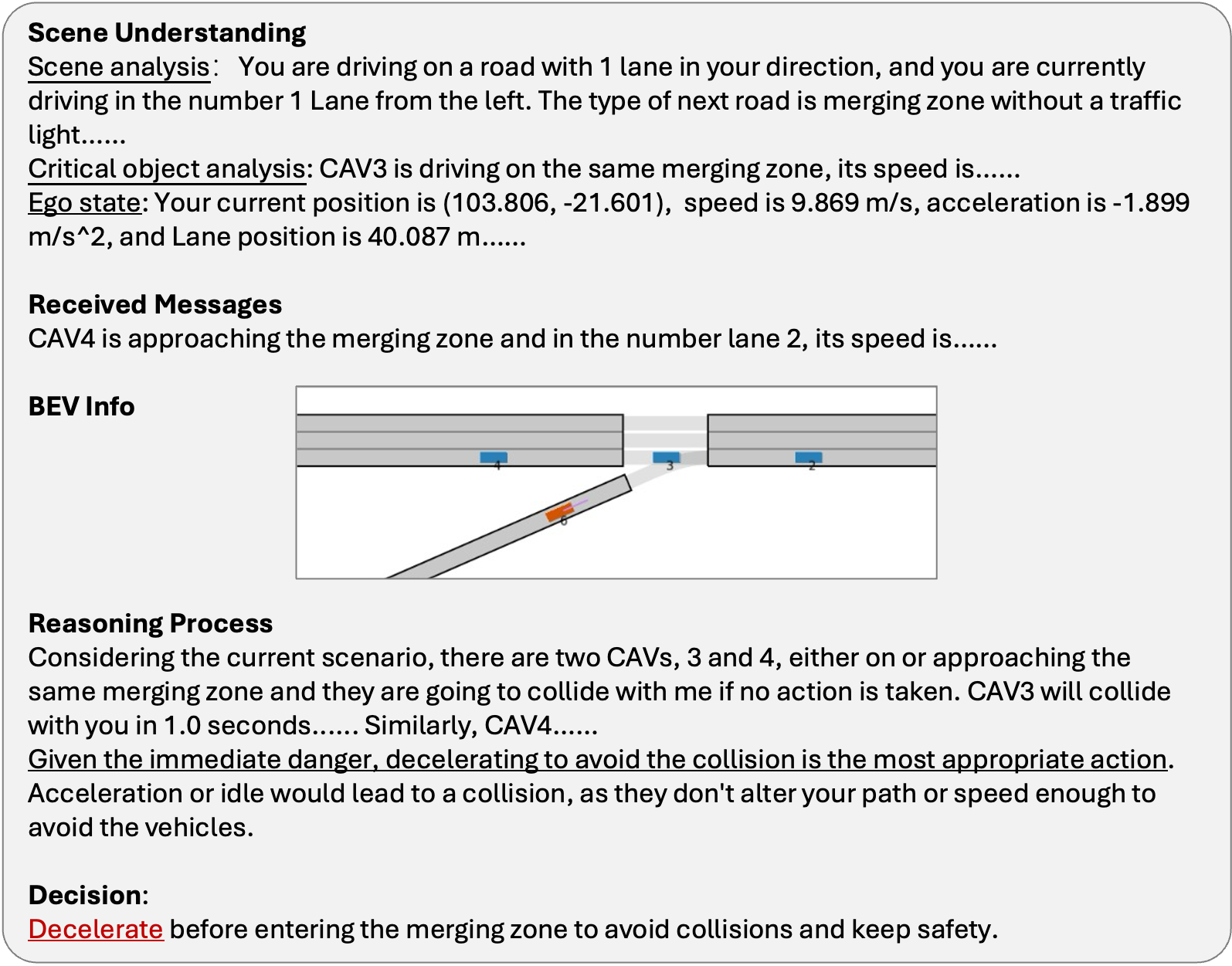}
    \vspace{-3mm}
    \caption{Qualitative analysis and prompts visualization.}
    \vspace{-5mm}
    \label{fig:qualitative-analysis}
\end{figure}

\textbf{Open-loop Evaluation.} To further demonstrate the effectiveness of our {AgentsCoMerge}, and its generalization ability to other scenarios, and real-world scenarios, we evaluate it on the nuScenes dataset. We compare our method with the following methods, NMP \cite{zengEndToEndInterpretableNeural2019}, SA-NMP \cite{zengEndToEndInterpretableNeural2019}, FF \cite{huSafeLocalMotion2021}, EO \cite{khuranaDifferentiableRaycastingSelfsupervised2022}, ST-P3 \cite{huSTP3EndtoEndVisionBased2022}, UniAD \cite{huPlanningorientedAutonomousDriving2023}, and GPT-Driver \cite{maoGPTDriverLearningDrive2023}. The results are shown in Table \ref{tab:open-loop}. 
Our method outperforms all baselines on both metrics. {AgentsCoMerge} achieves an average L2 error of 0.59m, which is 42.71\% lower than UniAD (best none-LLM method) and 29.76\% lower than GPT-Driver (best LLM method). For collision rate, our method achieves 0.29\% on average, with 0.00\% in the 1s prediction horizon. These results demonstrate both the effectiveness of our method and its strong generalization ability through hierarchical planning.
\textcolor{mycolor}{In addition, we evaluate the performance on HighD dataset \cite{8569552} with four other methods including CS-LSTM \cite{8575356}, Dual-Trans \cite{10061336}, WSiP \cite{Wang_Wang_Yan_Wang_2023}, and Social GAN \cite{8578338}. 
As shown in Table \ref{tab:rmse}, the overall average RMSE of 0.73m of our method represents a 21.5\% improvement over the next best baseline, validating our approach's effectiveness for trajectory prediction in real highway scenarios.}


\subsection{Ablation Study}
\label{sec:ablation}

In this section, we conduct an ablation study to investigate the effectiveness of each component in our proposed method.

\textbf{Effect of Inter-agent Communication.} In order to evaluate the effect of inter-agent communication, we conduct an ablation study to compare the performance of our proposed method with and without communication. The results are shown in Table \ref{tab:ablation-communication}. We can see that the communication mechanism can significantly improve the performance of our proposed method. Specifically, CS is improved by 5.75\%, ES is improved by 23.82\%, SS is improved by 10.51\%, and DS is improved by 11.08\%. The results demonstrate the necessity of the inter-agent communication, without communication, the ego vehicle's observation is limited to its own sensor data and may not notice the vehicles' state in the main road, leading to safety risks. In addition, without communication, the ego vehicle may not be able to negotiate with other vehicles the order to enter the main road, which may lead to inefficiency. The communication mechanism can help the ego vehicle obtain more information about the surrounding vehicles and make better decisions, which can improve the overall performance.

\textbf{Effect of Reinforcement Reflection Guided Training.} 
In order to evaluate the effect of the reinforcement reflection guided training, in this section, we conduct an ablation study to compare the performance of our proposed method with and without the reinforcement reflection guided training. As shown in Table \ref{tab:ablation-rrgt}, the results demonstrate that the reinforcement reflection guided training can significantly improve the performance of our proposed method. Specifically, the L2 error is reduced by 90.99\% and the collision rate is reduced by 97.67\% on average. Without the reinforcement reflection guided training, the average L2 error is 6.55 m, which is unacceptable for an autonomous driving system, and so it fails to predict the future trajectory accurately. For the collision rate, the system also fails to prevent collisions with other objects in the scene. These phenomena demonstrate the indispensable need of reinforcement reflection guided training and show its effectiveness for improving the performance of our proposed method.

\begin{figure}[t]
    \centering
    \includegraphics[width=1\linewidth]{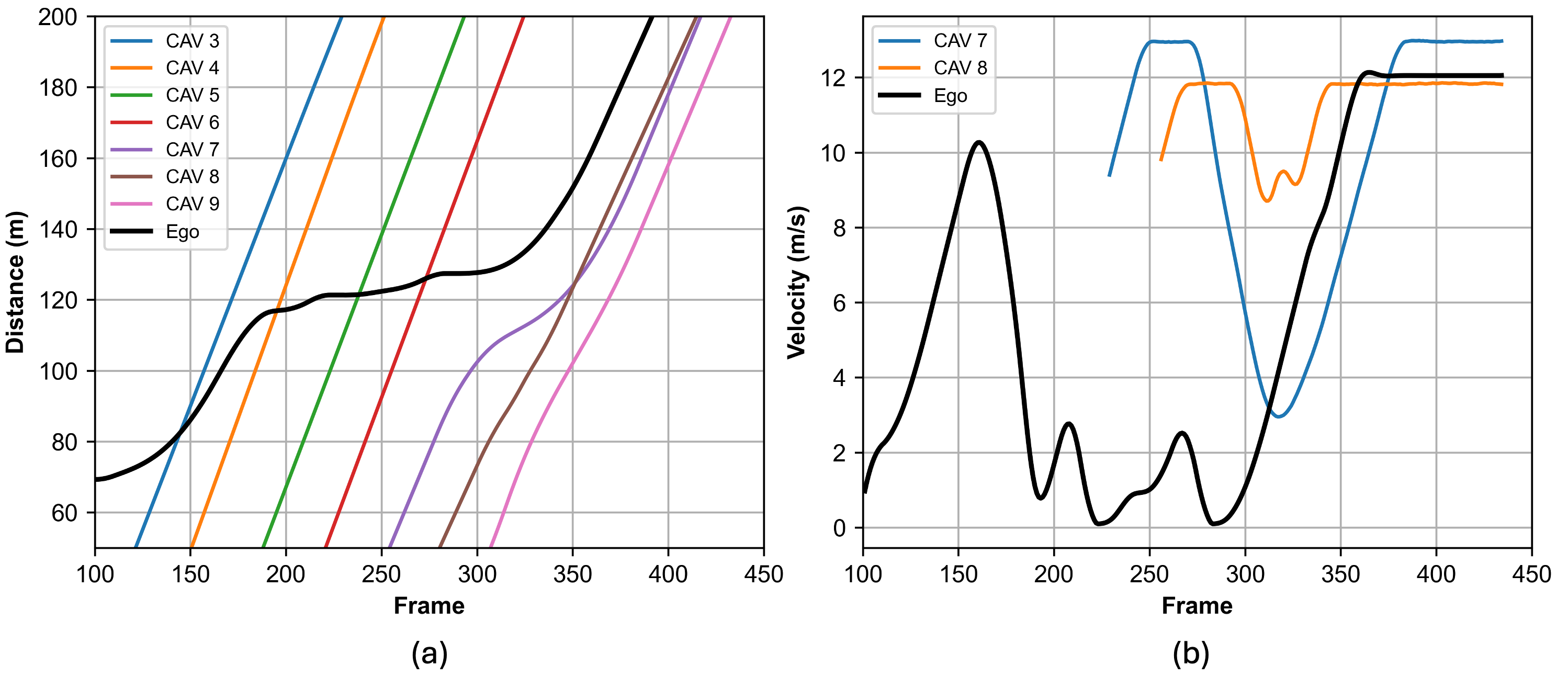}
    \vspace{-8mm}
    \caption{Results of the travel distance (m) and the velocity (m/s) of the ego CAV and other CAVs. In both subfigures, the bold black line represents the results of the ego CAV.}
    \label{fig:case-study}
    \vspace{-5mm}
\end{figure}

\subsection{Case Study}

In this section, we provide a case study to further enhance the understanding of our proposed method. As shown in Fig. \ref{fig:qualitative-analysis}, we visualize the reasoning process of an agent. We can see that the scene analysis and critical object analysis provide effective information about the surrounding environment. In addition, CAV4 is in the collaborative area and approaching the merging zone, so it passes its state information to the ego CAV to help the ego CAV make better decisions. In this case, without communication, the ego CAV may not know the approaching of CAV4 and just keep speed up to merge into the main road,
which may result in a potential collision. With all the analysis and messages, the ego CAV analyzes the information and decides to yield to CAV3 and CAV4 and decelerate.
This example demonstrates the effectiveness of our proposed method and shows the strong reasoning ability of an agent to make decisions in complex scenarios.

In addition, we also plot the travel distance and velocity of the ego CAV and other CAVs in Fig. \ref{fig:case-study}. We can see that the ego CAV enters the main road after CAV6 and before CAV7. Before merging into the main road, the ego CAV just decelerates and yields to the other vehicles, which can avoid collisions and ensure safety. After merging into the main road, the ego CAV accelerates to the speed range of the surrounding vehicles and maintains a safe distance from other vehicles. CAV7 and CAV8 in the main road also decelerate to give space to the ego CAV, which can ensure the smooth merging of the ego CAV. In addition, as shown in Fig. \ref{fig:case-study}(a) the curve of CAV7 has an intersection with the curve of CAV8, this is because CAV8 just changes the lane and overtakes CAV7. 

Overall, this case study further demonstrates the states of CAVs both on the ramp and the main road, and shows their interactions among each other, which shows that our proposed method can make reasonable decisions and ensure the safety and efficiency of the merging process.
\section{Limitations and Future Work}


\textcolor{mycolor}{In this section, we discuss the limitations of our proposed framework. We acknowledge that our current implementation using LLMs faces some computational challenges in terms of computational cost and real-time deployment. The LLM models have billions of parameters, leading to high computational resource requirements and an average decision time of around 10 seconds per round. While this current decision latency is not suitable for real-time autonomous driving, this limitation is primarily technical rather than fundamental. Recent advances show promising directions for optimization through model compression and acceleration techniques \cite{10.5555/3600270.3601459, 10.5555/3618408.3619327}, as well as quantization and tensor operator fusion optimization \cite{MLSYS2024_42a452cb, 10.1145/3520142}. These optimizations have successfully reduced decision times from seconds to milliseconds. The practical viability of LLM-based approaches has been demonstrated by some automotive companies (e.g., Li Auto) that have successfully deployed LLM-based decision systems in autonomous vehicles \cite{tian2024drivevlmconvergenceautonomousdriving}. While computational optimization is important, it is out of the scope of this paper. Our primary contribution is the design of a multi-agent collaborative decision-making framework that improves autonomous vehicles' ability to make driving and coordinated decisions. The computational aspects can be addressed through ongoing optimization research.}
\vspace{-3mm}

\section{Conclusion}
\label{sec:conclusion}

In this paper, we have presented a novel framework AgentsCoMerge to tackle the collaborative decision-making problem at multi-lane merging zone. The proposed approach leverages large language models, which possess remarkable capabilities that closely resemble human thought processes. AgentsCoMerge integrates a scene understanding module, which can not only understand text information but also can process visual information. Following the scene understanding module, a hierarchical planning module is introduced to generate the decision and plan the future waypoints. In addition, to facilitate an agent's ability to communicate with others and make collaborative decisions, an inter-agent communication module is developed. Finally, we design a reinforcement reflection guided training strategy to train the agents, which can further enhance the agents' decision-making capabilities. Extensive experiments have been conducted as part of this study and the superior performance of our proposed method has been affirmed.


%

\bibliographystyle{IEEEtran}
{
        \footnotesize
        \bibliography{ref,ref2,ref3}
        }

\end{document}